\documentclass[journal]{IEEEtran}

\usepackage{cite}
\usepackage{epsfig}
\usepackage{graphicx}
\usepackage{caption}
\usepackage{subcaption}
\usepackage{multirow}
\usepackage{booktabs, makecell, multirow}
\usepackage{amsmath}
\usepackage{url}
\usepackage{comment}
\newcommand*{\email}[1]{\texttt{#1}}

\begin{document}
\title{Scalable and Real-time Multi-Camera Vehicle Detection, Re-Identification, and Tracking}

\author{Pirazh Khorramshahi$^{1}$, Vineet Shenoy$^{1}$, Michael Pack$^{2}$, Rama Chellappa$^{1}$ \\
\vspace{0.3cm}
$^{1}$Artificial Intelligence for Engineering and Medicine Lab, Johns Hopkins University, Baltimore, MD \\
$^{2}$Center for Advanced Transportation Technology Laboratory, University of Maryland, College Park, MD \\

\email{\tt\small \{pkhorra1, vshenoy4, rchella4\}@jhu.edu, packml@umd.edu}}

%\thanks{}

\maketitle

\begin{abstract}
Multi-camera vehicle tracking is one of the most complicated tasks in Computer Vision as it involves distinct tasks including Vehicle Detection, Tracking, and Re-identification. Despite the challenges, multi-camera vehicle tracking has immense potential in transportation applications including speed, volume, origin-destination (O-D), and routing data generation. Several recent works have addressed the multi-camera tracking problem. However, most of the effort has gone towards improving accuracy on high-quality benchmark datasets while disregarding lower camera resolutions, compression artifacts and the overwhelming amount of computational power and time needed to carry out this task on its edge and thus making it prohibitive for large-scale and real-time deployment. Therefore, in this work we shed light on practical issues that should be addressed for the design of a multi-camera tracking system to provide actionable and timely insights. Moreover, we propose a real-time city-scale multi-camera vehicle tracking system that compares favorably to computationally intensive alternatives and handles real-world, low-resolution CCTV instead of idealized and curated video streams. To show its effectiveness, in addition to integration into the Regional Integrated Transportation Information System (RITIS) \footnote{\url{https://www.ritis.org}}, we participated in the 2021 NVIDIA AI City multi-camera tracking challenge and our method is ranked among the top five performers on the public leaderboard.
\end{abstract}

\begin{IEEEkeywords}
Vehicle Detection, Re-Identification, Single Camera Tracking, Multi Camera Tracking, Domain Adaptation, Real-Time, Scalable.
\end{IEEEkeywords}

% \fi
\IEEEpeerreviewmaketitle
\section{Introduction}
Multi-Camera Tracking (MCT) is the task of tracking an unknown number of objects across a number of mounted cameras that may or may not have overlap in their fields-of-view. This makes MCT to be one of the most complicated tasks in Computer Vision as it involves several fundamental vision tasks, namely Object Detection, Tracking, and Re-identification. Due to recent advancements in Computer Vision, thanks to Deep Learning and Deep Convolutional Neural Networks (DCNN) in particular, interests in high performance MCT systems have been rapidly growing. The underlying reason for this rapid growth is that MCT has great applications in intelligent transportation systems. 

Transportation operations have historically relied on relatively expensive and hard to maintain Bluetooth and Wi-Fi re-identification sensors for understanding the flow, routing, and origins and destinations (O-D) of traffic throughout urban environments. While agencies are now able to purchase O-D datasets from location-based services companies, the penetration rates are still relatively low, latency is high, and proposed legislative actions and new cell phone privacy policies may threaten the proliferation of this form of data collection. However, nearly every major city has a traffic operations program that has deployed hundreds or even thousands of CCTV cameras. These cameras, when equipped with MCT, could prove to be a new, higher quality source of real-time and archived O-D and routing data for the transportation planning and operations community. Furthermore, MCT could also support other real-time safety applications related to wrong-way driving detection, event detection, and other erratic and unsafe driving behavior which contribute to increased safety and security. 

In this paper, we focus on developing an MCT system tailored for vehicles that can operate in real-time. As mentioned above, the MCT system involves three distinct vision tasks. 1- \textbf{Vehicle Detection:} Detection is responsible for localizing vehicles of various types at different locations and scales within the camera view. High quality detections are critical to the success of the MCT system as it impacts the performance of all the downstream modules. 2- \textbf{Single Camera Vehicle Tracking}: Upon receiving detections, a multi-object tracker attempts to associate the detected bounding boxes belonging to individual identities simultaneously and predict their future locations while being robust to occlusion and variations in velocity of vehicles. Note that there are methods to perform multi-object detection and tracking via a single model\cite{centertrack,tracktor_2019_ICCV}. However in our work we find that having separate modules for the two tasks helps us to identify potential issues and optimize each to the MCT task. 3- \textbf{Vehicle Re-identification:} Re-identification aims to obtain discriminative appearance embeddings from the tracked vehicles in each camera so that we can associate different single camera tracks corresponding to individual identities. The extracted visual embeddings should be robust to variations in orientation and lighting conditions that may be different from camera to camera. Once single camera tracks and their corresponding representations have been computed for all the cameras, a clustering algorithm is needed to associate single camera tracks to unique identities based on the computed visual features and spatio-temporal information that comes naturally with each single camera track. Since the number of true identities is not known beforehand, the clustering algorithm should be independent of the number of vehicle identities to perform this task.

There have been several works that address MCT for vehicles \cite{Liu_2021_CVPR,wu2021multi,ye2021robust,he2019multi}. However, all these works only attempt to maximize the multi-camera tracking accuracy on benchmark datasets without any consideration for the inference time and computational complexity. As a result, they require significant amount of computation time and resources to process a number of videos with the duration of only few minutes. In contrast, in this work, we present an MCT system that can run in real-time and provide timely results for any downstream goals. To demonstrate the effectiveness of our method, we integrate our MCT system, as a prototype, in the RITIS system which is a data-driven platform from the University of Maryland for transportation analysis, monitoring, and data visualization. RITIS has access to the real-time traffic camera feeds that are provided by state and local departments of transportation traffic centers to the University of Maryland via electronic data feeds from traffic management centers. Our MCT system provides real-time multi-camera capability which is useful for a variety of transportation applications. In addition, we evaluated our system on the 2021 NVIDIA AI City Multi-Camera Vehicle Tracking challenge and were ranked among the top five competitors.

The rest of the paper is organized as follows. In section \ref{sec:RelatedWork}, we review recent works on object detection, multi-object tracking, and vehicle re-identification.  The detailed architecture of our multi-camera tracking pipeline is discussed in section \ref{sec:Method}. Next, in section \ref{sec:Experiments}, we discuss the implementation details, validation data and its statistics, and evaluate our approach on real-time traffic data as well the Multi-Camera Vehicle Tracking challenge of the 2021 AI City challenge to demonstrate its effectiveness and validate our design choices. Finally we conclude in section \ref{sec:conclusion}.

\section{Related Work}\label{sec:RelatedWork}
Here we review most relevant works on vehicle detection, tracking, and re-identification as these are the pillars of the multi-camera vehicle tracking task.

\textbf{Object Detection:} 
Yolo \cite{Redmon_2016_CVPR} and it variants\cite{redmon2017yolo9000,redmon2018yolov3,bochkovskiy2020yolov4}, Single Shot\cite{liu2016ssd}, RetinaNet\cite{lin2018focal}, Faster R-CNN\cite{fasterRcnn} and Mask R-CNN\cite{he2017mask} are popular choices of object detectors to be employed in variety of applications. Many previous works use these models as off-the-shelf detectors trained on the large-scale object detection COCO dataset \cite{coco}. Moreover, RetinaNet, Faster R-CNN and Mask R-CNN are particularly popular as they are regularly maintained by Detectron library \cite{wu2019detectron2} which significantly facilitates their adoption by researchers. More recently, EfficientDet \cite{tan2020efficientdet} that has a weighted bi-directional feature pyramid network to allow for easy and fast multi-scale feature fusion has been developed. Many works fine-tune these detectors on external vehicle data such as the UA-DETRAC \cite{CVIU_UA-DETRAC} dataset. One extension of this work is SpotNet \cite{spotnet} which uses attention mechanisms to locate roads and driving surfaces, and limiting detections to these surfaces. Similarly, the authors in \cite{fgbrnet} propose FG-BR Net which focuses on objects in the foreground. In the first of this two-stage method, high quality regions of interest (RoI) proposal are fed to the detector by suppressing background features while amplifying feature activations in foreground objects. The second stage, assuming that there are errors in the first stage, refines the first stage proposals with pairwise non-weighted local background fusions. Some methods from face detection have found their way into vehicle detection. The authors in \cite{facedetection} first use a region proposal network to extract RoI proposals, which is followed by an adaptively-generated Gaussian kernel which extracts local features. The output of this stage is fed to an LSTM \cite{lstm} module to encode global context, and subsequently to a classifier and bounding box regression module. Concepts tested on pedestrian-detection datasets have also found their way into vehicle detections, with the authors in \cite{pedestriandetection} proposing a convolutional neural network that predicts centers and scales of bounding boxes. This method obviates the need for anchor boxes and avoids computationally-heavy post processing steps usually involved with key-point pairing-based detectors. 

\textbf{Multi-Target Single Camera Tracking:} Most tracking methods first detect and then associate objects. The local methods \cite{localmethod1,localmethod2,localmethod3} consider only two frames at a time, but do not perform well when occlusion, pose variations, and camera motion are present. In contrast \textit{global} methods consider multiple frames concurrently and solve network flow problems \cite{globalmethod1,globalmethod2,globalmethod3}. One of the more recent trackers is the Deep Affinity Model \cite{affinitymodels}, which performs detection and data association concurrently. As input, two frames of a video (not necessarily consecutive) are fed through two networks with shared parameters and object features are extracted.% Affinities are computed through and exhaustive permutation of the features, and the network is trained with a novel loss function for the task. 
Another recent local method is CenterTrack \cite{centertrack}, which represents all objects as a point at the center of a bounding box. To associate detections across frames, the distance between the object center in the previous frame and the predicted offset in the current frame is calculated. A new object identifier is created when there is no previous object center within a certain radius of the current object center. Similar to CenterTrack, \cite{tracktor_2019_ICCV} associates detections across frames through an offset under the assumption that motion between frames is small; the authors regress the bounding box location from the previous frame to the current frame using the regression head of a Faster R-CNN detector \cite{fasterRcnn} and the deep features in the current frame. If the predicted bounding box after regression has a high Intersection-over-Union (IoU) with an incoming bounding box, then the track has been estimated successfully. However we found these end-to-end models to be burdensome to modify and adapt to new data domains in addition of being computationally expensive for a multi-camera tracking system. A successful example of an online tracking model which relies on pre-computed detections, is SORT\cite{bewley2016simple} and its enhanced version with deep associations, namely DeepSORT \cite{wojke2017simple}. These trackers benefit from a linear state-space model that approximates dynamics of targets and are suitable choices for real-time applications. 

\textbf{Vehicle Re-Identification:}
Most of the recent successful works in vehicle re-identification have benefited from attention models to compute robust representations \cite{li2021self,khorramshahi2020devil,peri2020towards,chen2020orientation,meng2020parsing,bai2020disentangled,khorramshahi2019dual,khorramshahi2019attention} for distinguishing vehicle identities and extracting minute details in vehicle images. Authors in \cite{khorramshahi2019attention,khorramshahi2019dual} incorporated attention maps by localizing pre-defined key-points to enhance re-identification in a supervised manner. Typically, extracted visual representation of vehicles are biased toward the orientation in which images are captured. To alleviate this issue, authors of \cite{meng2020parsing} and \cite{bai2020disentangled} propose to learn view-aware aligned features and to disentangle the orientation from visual features respectively. In addition, \cite{zhao2021heterogeneous} proposes a heterogeneous relational model to extract region-specific features and incorporate them based on their relation into a unified representation. To overcome the demand to collect expensive annotations for learning distinguished vehicle parts, \cite{khorramshahi2020devil, peri2020towards} propose to learn salient regions of vehicles which encode identity-dependant information, in a self-supervised manner via the task of residual learning. \cite{li2021self} attempts to achieve the same goal via the pretext task of image rotation and degree prediction to encode geometric features along with the global appearance. Finally, authors in \cite{zhao2021phd} emphasize the importance of video-based approach as opposed to the image-based approach for vehicle re-identification and introduced the VVeRI-901 dataset to contribute to this research direction.

\begin{figure}
    \centering
    \includegraphics[width=0.49\textwidth]{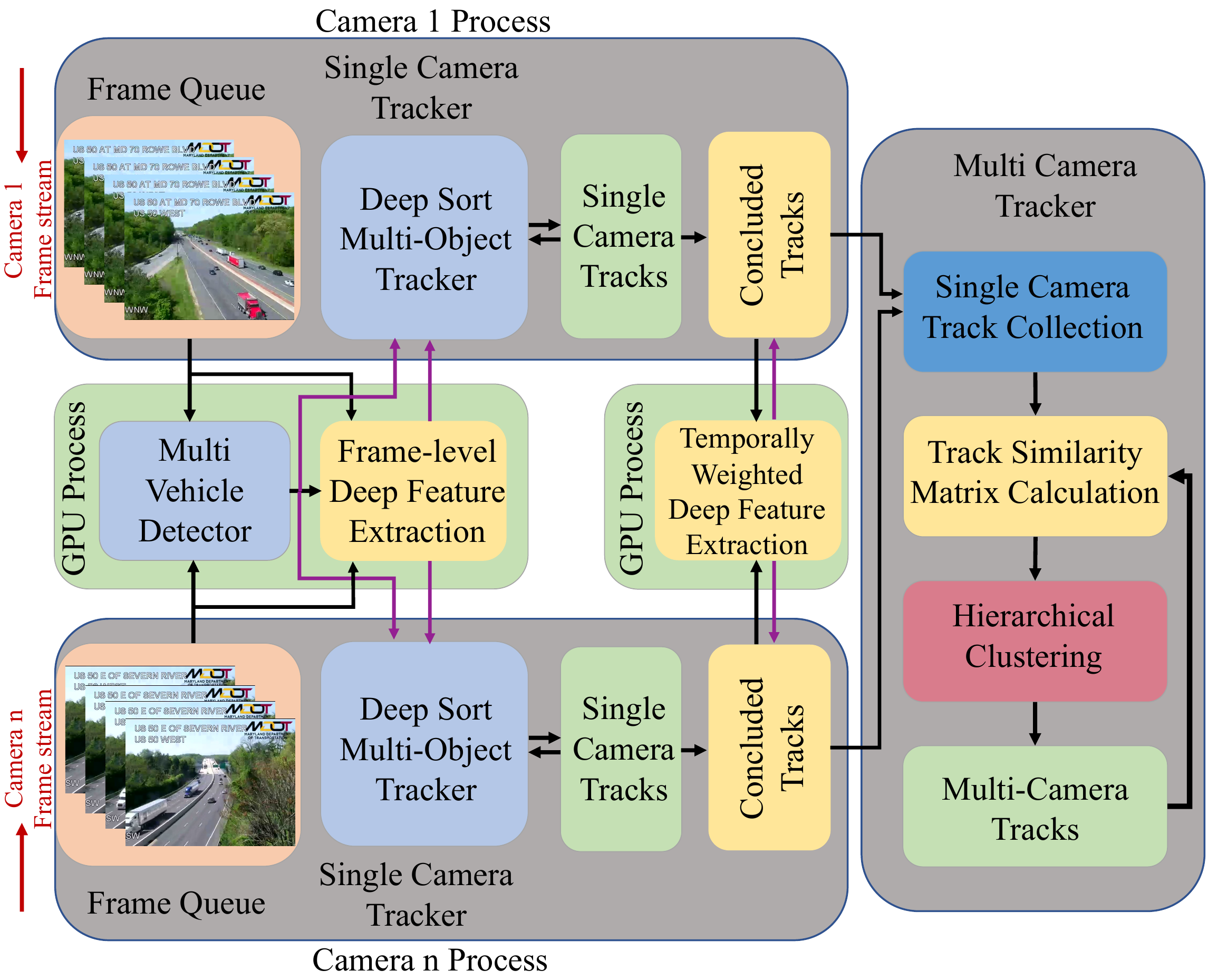}
    \caption{Multi-Camera Vehicle Tracking in Real-Time Pipeline. For each camera in the network, a process gathers frames and sends them to a GPU process for detection and frame-level feature extraction. Next, the bounding boxes and corresponding deep features are sent back to the respective camera process for single camera tracking. Once a single camera track is concluded, all the frame-level deep features are summarized into a single representation. Finally,  concluded tracks are collected and their similarity is measured to updates multi-camera tracks.}
    \label{fig:MTMCT_pipeline}
\end{figure}

\section{Method}\label{sec:Method}
The overview of our proposed pipeline for real-time multi-camera vehicle tracking is shown in Figure \ref{fig:MTMCT_pipeline}. It consists of several modules which are explained in the following sections.  

\subsection{Camera Process}
In our system, we designed a camera-specific process for each of the $n$ cameras that are in the network so that we can process each camera in parallel and in real-time. These processes are responsible for receiving frames from each video stream and storing them in the frame queues. Upon receiving a frame it is sent to the GPU process so that vehicles can be localized and their corresponding frame-level deep appearance features can be calculated. Subsequently, they are returned to their respective camera process where the single camera tracker initiates new tracks or continues tracking previously tracked vehicles depending on the matching criteria which will be explained in section \ref{subsubsec:sct_method}. 

\begin{figure}[b]
     \begin{subfigure}{0.2\textwidth}
        \includegraphics[width=\textwidth]{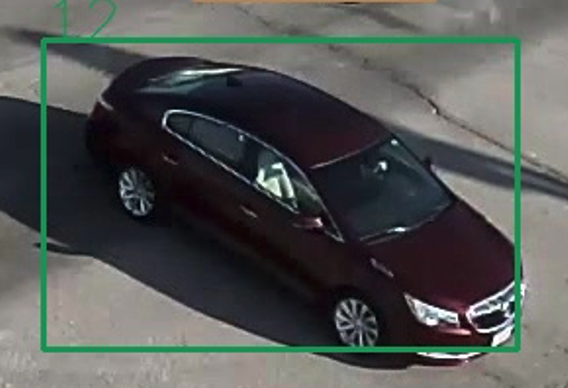}
        \caption{Original}
     \end{subfigure}
     ~
     \centering
     \begin{subfigure}{0.18\textwidth}
        \includegraphics[width=\textwidth]{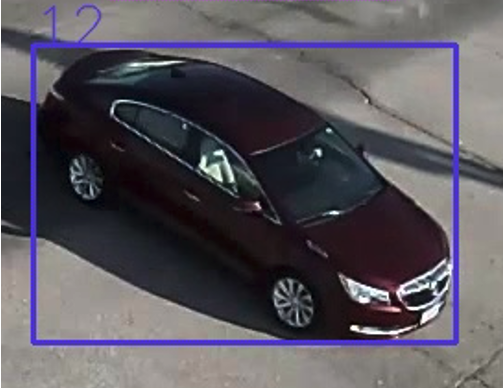}
        \caption{Modified}
     \end{subfigure}
     \caption{Impact of replacing the Kalman filter's predicted state variables for the current frame with the observations from the matched detection. This improves the tightness of bounding boxes around vehicle tracks and quality of extracted features.}
     \label{fig:DeepSort_Modificiation}
 \end{figure}

\subsubsection{Single Camera Tracker} \label{subsubsec:sct_method}
As we aim to design a real-time MCT system, we need to make sure all the involved components are fast and efficient while maintaining high accuracy. In addition, it is paramount that the single camera tracker has a small number of ID switches as the number of comparisons in the similarity matrix computation (required for solving multi-camera tracking) grows quadratically.  Therefore, we choose DeepSort \cite{wojke2017simple} as our single camera tracker. DeepSort is the successor of SORT \cite{bewley2016simple} which is a very lightweight and simple multi-object tracking based on the Intersection over Union (IoU) criteria. To make tracking more robust and resilient to ID switches, DeepSort incorporates appearance information. This tracker approximates the dynamics of each target vehicle with a linear state space model. In the original implementation, the state space is defined as the following vector:
\begin{equation*}
    {[u,v,r,h,\dot{u},\dot{v},\dot{r},\dot{h}]}^T
\end{equation*}
 where $u$, $v$, $r$, and $h$ are the bounding box's center horizontal and vertical coordinates, aspect ratio, and height respectively. In addition, their time derivatives are also included as the state space variables. However, we modify $u, v$ to be the center point of the bottom edge of a bounding box as this point is closer to the surface of the road compared to the center of the bounding box and results in smaller distortion due to the missing depth information. DeepSort propagates the state space distribution to the current time step using a Kalman filter prediction step and obtains the predicted observation vector $[\hat{u},\hat{v},\hat{r},\hat{h}]$ for a track's state at current time. However, we modify this so that whenever there is a matched detection to a predicted track's state, we use the information from the matched detection as the track's current state. Figure \ref{fig:DeepSort_Modificiation} shows the impact of this modification. This modifications is mainly due to the fact that bounding box locations from the detection module are generally more accurate compared to predictions from the tracker as it has assumptions on vehicles motion. As a vehicle track is concluded, all the extracted frame-level deep appearance features for its life span are temporally sorted and sent to a GPU process to be temporally weighted and summed. Therefore, a single representation vector is obtained to represent that vehicle track. The concluded track, its appearance representation, and meta-information are collected by the multi-camera tracker that is responsible for identifying multi-camera vehicle tracks. 
 
\begin{figure}[t]
     \begin{subfigure}{0.23\textwidth}
        \includegraphics[width=\textwidth]{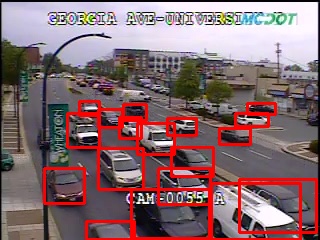}
        \caption{RITIS Platform}
     \end{subfigure}
     ~
     \centering
     \begin{subfigure}{0.2\textwidth}
        \includegraphics[width=\textwidth,height=0.85\textwidth]{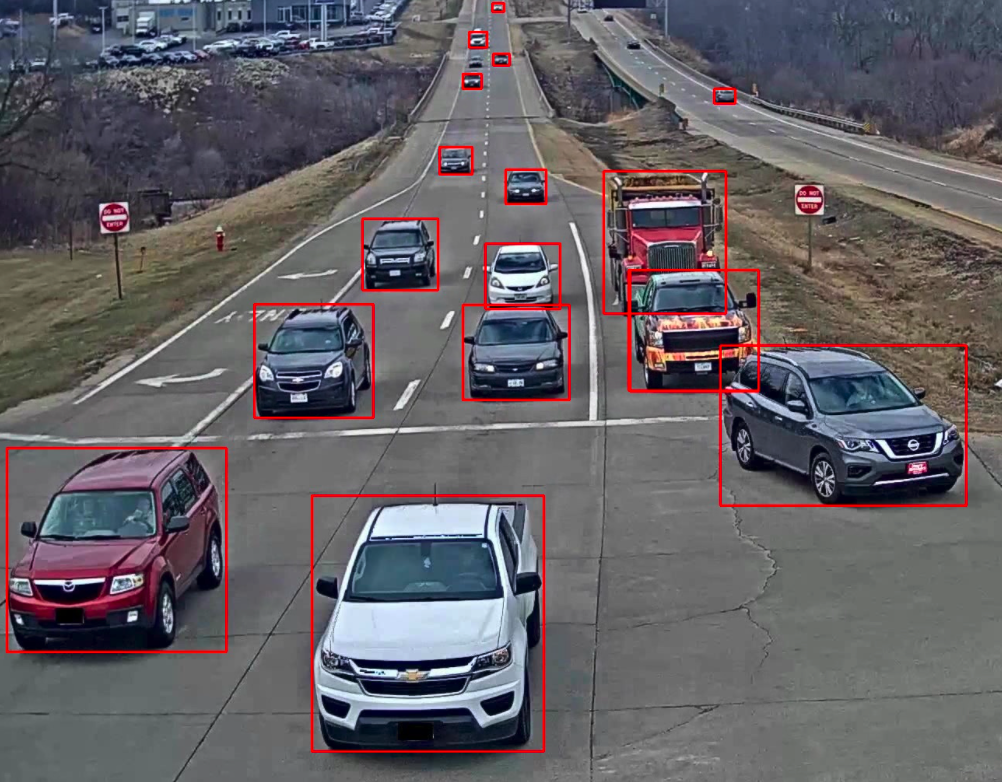}
        \caption{AI City Challenge}
     \end{subfigure}
     \caption{Vehicle detection Results on sample frames after filtering out low-confidence boxes and applying NMS.}
     \label{fig:sample_det}
\end{figure}
 
 \subsection{GPU Process}
 Deep learning technology owes its success to the development of Graphical Processing Units (GPU). Almost all the recent successful Computer Vision methods benefit from GPUs. Our approach follows this trend and we run our vehicle detection and re-identification models in GPU accessible processes. 
 
 \subsubsection{Multi-Vehicle Detection}
 To localize vehicles with high certainty, CNN-based object detectors are desirable candidates as they have achieved state-of-the-art results across different benchmarks and run efficiently on GPU cards. In this work, we choose Faster R-CNN \cite{fasterRcnn}, RetinaNet \cite{lin2018focal}, and EfficientDet \cite{tan2020efficientdet} to account for different scenarios, image resolution, and latency. Given a video frame $I$, the vehicle detector returns a list of detections in the form of 
\begin{equation*}
    {[x_1,y_1,x_2,y_2,\alpha, \beta]}
\end{equation*}
where $(x_1,y_1)$ and $(x_2,y_2)$ show the top-left and bottom-right points of the bounding box encompassing a vehicle while $\alpha$ and $\beta$ represent the detection confidence and class label of the detected box, respectively. Note that a minimum value of confidence score $\alpha_{min}$ is used to filter out unreliable detections. In addition, Non-Maximal Suppression (NMS) ensures that duplicate detections (with high IoU) for a single vehicle are merged into a single box. However, in extreme cases of occlusion this may remove the box for the occluded vehicle. While this may seem to have a negative impact and interrupts the tracking process, it precludes the extraction of erroneous features for highly occluded vehicles. Also as argued in \cite{wojke2017simple}, Deep Sort can recover single camera tracks that are broken up to a certain time window. Figure \ref{fig:sample_det} shows detection results for frames randomly sampled from RITIS and AI City challenge data. It is worth mentioning that, on GPU cards, object detectors have the batch processing capability. At each time step, the incoming frames from all camera processes are put into a batch and detection results for all current frames of different cameras are computed at once. 

\subsubsection{Vehicle Re-Identification}\label{subsubsec:vereid_method}
\begin{figure}
    \centering
    \includegraphics[width=0.5\textwidth]{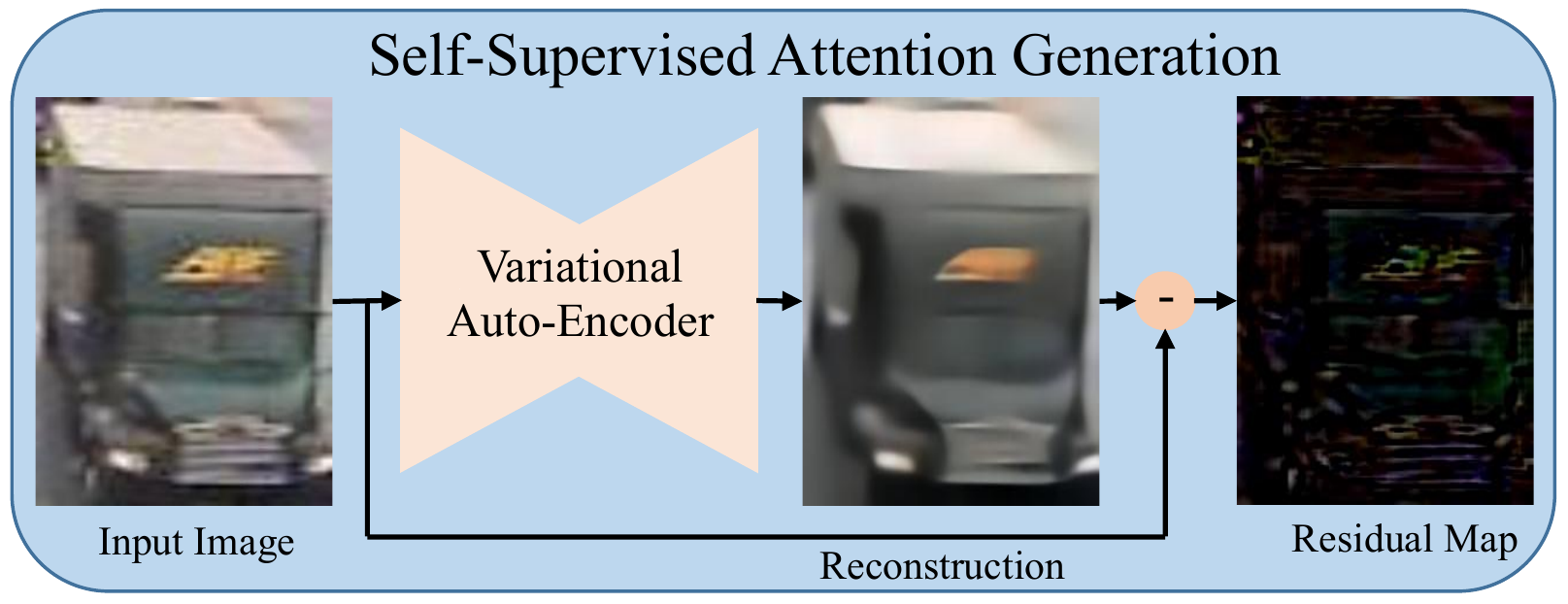}
    \caption{The input vehicle image is coarsely reconstructed to remove the small-scale details. By subtracting the reconstruction from the input image the residual map is obtained that captures high-level details such as logos and grill design as shown above and can serve as an attention map to excite intermediate feature maps in a CNN.}
    \label{fig:VAE}
\end{figure}

Extracting discriminative deep features is a critical piece of a multi-camera tracking system. Small inter-class variation (different vehicles can appear very similar and share the same make, model and color) and large intra-class variation (a vehicles appearance can drastically change under different viewpoints) make vehicle re-identification challenging. Therefore, to increase the discrimination power, attention-based vehicle re-id models have been developed to incorporate details from local regions that are unique to vehicle identities. However, most of these models increased re-id accuracy at the expense of increased computational resources and inference time. Therefore, in this work, we employ the Excited Vehicle Re-identification (EVER) \cite{peri2020towards}, a state-of-the art vehicle re-identification model and a top performer in 2020 and 2021 City-scale Multi-camera Vehicle Re-identification challenge\cite{Naphade20AIC20,Naphade21AIC21}, that benefits from the self-supervised attention generation mechanism introduced in \cite{khorramshahi2020devil} without adding any overhead to the inference time. During training, intermediate layers of EVER are excited via the residual maps generated by a conditional variational autoencoder network in a self-supervised manner as proposed in  \cite{khorramshahi2020devil}. The residual maps serve as pseudo-saliency maps highlighting small-scale details in vehicle images. Figure \ref{fig:VAE} shows an example of how such residual maps are generated. As training progresses, the intensity of excitation $\gamma$ reduces as a cosine function of training epoch, \emph{i.e.} $\gamma(m) = 0.5 \times \left(1 + \cos(\frac{\pi m}{M})\right)$ where $m$, and $M$ are current epoch and total number of training epochs.
As a result, once training is finished, the inference only involves a single forward pass of a ResNet\_IBN-a \cite{pan2018two} which is quite efficient and fast, a desirable property for a real-time multi-camera tracking system. Note that we use the ResNet\_IBN-a as the backbone architecture for EVER model since it is shown to be a superior candidate for the task of re-identification \cite{he2020fastreid}.

\begin{figure}
    \centering
    \includegraphics[width=0.45\textwidth]{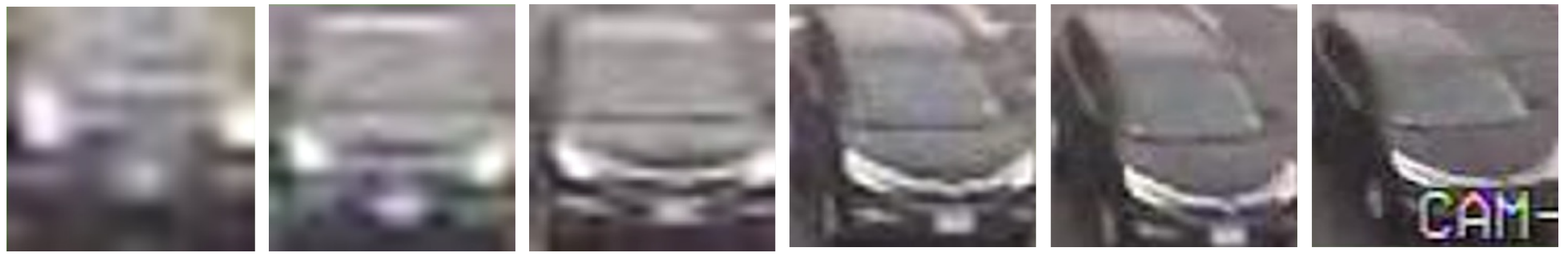}
    \caption{A Vehicle track images ordered in time from left to right.}
    \label{fig:track_feat}
\end{figure}

\begin{figure}
    \centering
    \includegraphics[width=0.49\textwidth]{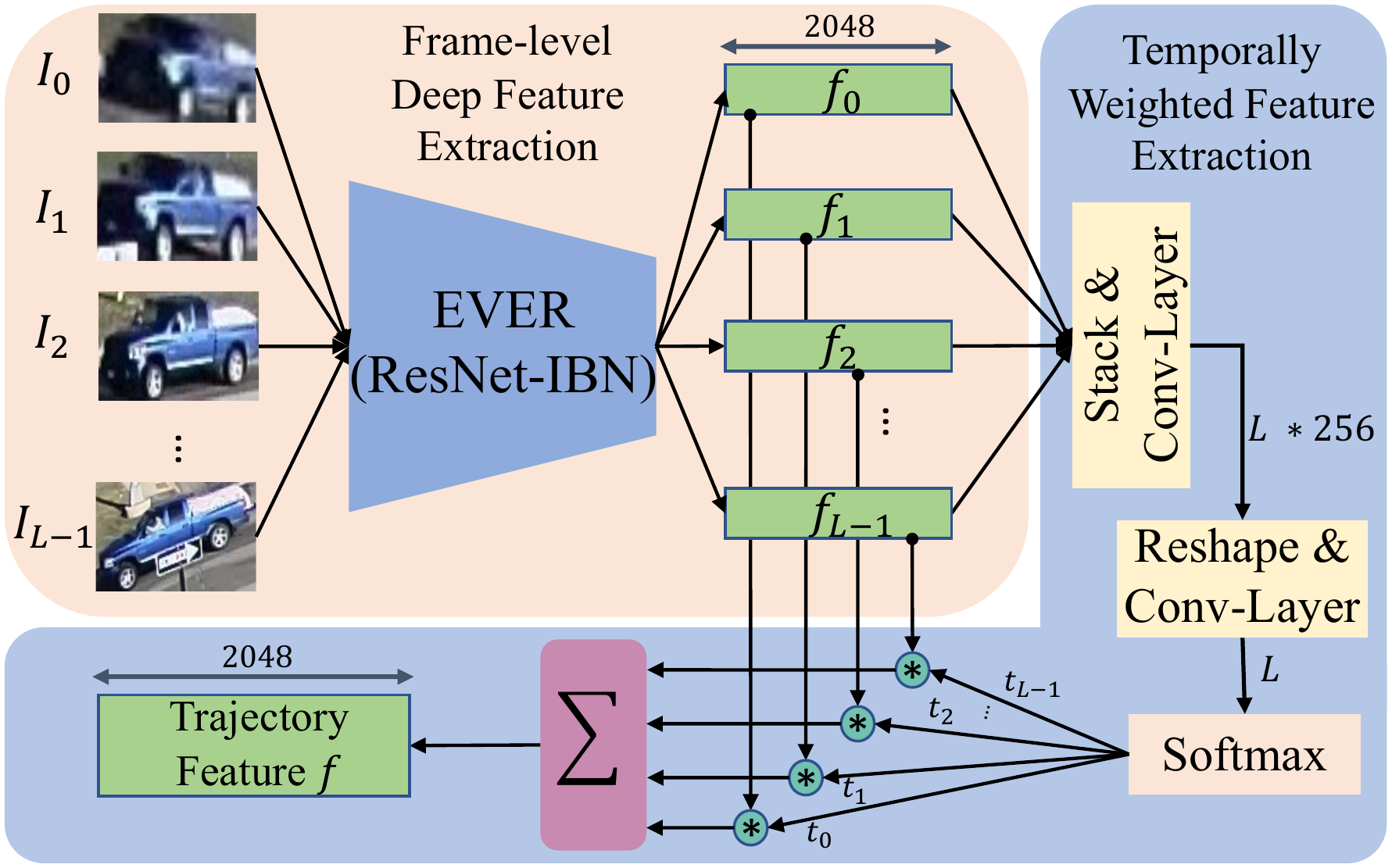}
    \caption{Computing vehicle track representation. In Frame-level Deep Feature Extraction stage, $2048$-dimensional features are computed in the GPU process. Once a single camera track of length $L$ is concluded, temporally-ordered frame-level features are sent to another GPU process to be weighted and averaged. In this step, all the $L$ frame-level features are stacked and passed through two convolutional layers with non-linearities to obtain $L$ scalar values corresponding to $L$ frames. Finally, the $L$ frame-level features are weighted by the softmax of $L$ scalar values and then summed to output the $2048$-dimensional trajectory representation $f$.}
    \label{fig:deep_feat}
\end{figure}

In contrast to the typical task of vehicle re-identification which aims to compare a query image against gallery images, in a real-world multi-camera tracking scenario, the task involves comparing a query track against gallery tracks. This requires computing a representation that summarizes the extracted frame-level features of a vehicle in an effective manner. Figure \ref{fig:track_feat} shows frames of a track and how they change as the vehicle passes by the camera. It shows that simply averaging all the extracted frame-level features can potentially ignore those frame-level representations that carry discriminative information on vehicle identity. For instance, when a vehicle is far from a camera, only large-scale information can be captured compared to the time it is close to camera. Since vehicle's displacement is smaller relative to when it is closer to the camera, most of the extracted frame-level features for the vehicle track only contain large-scale information and small-scale information that is critical for successful re-id is ignored. Therefore, it is important to learn how to effectively fuse frame-level information into a single representation also know as video-based re-identification. To solve this issue, our multi-camera tracking system is equipped with a Temporally Weighted Deep Feature Extractor module that is inspired by the work of \cite{gao2018revisiting} to compute a single representation vector for the entire length of a vehicle track. Figure \ref{fig:deep_feat} shows the mechanism of extracting frame-level deep features and how they are temporally weighted and averaged to obtain a single representation for the entire life span of a vehicle track. In addition, this summarization significantly simplifies the subsequent feature matching to obtain multi-camera vehicle tracks and is less memory demanding. 

\begin{figure}[t]
    \centering
     \begin{subfigure}{0.45\textwidth}
        \includegraphics[width=\textwidth,height=0.45\textwidth]{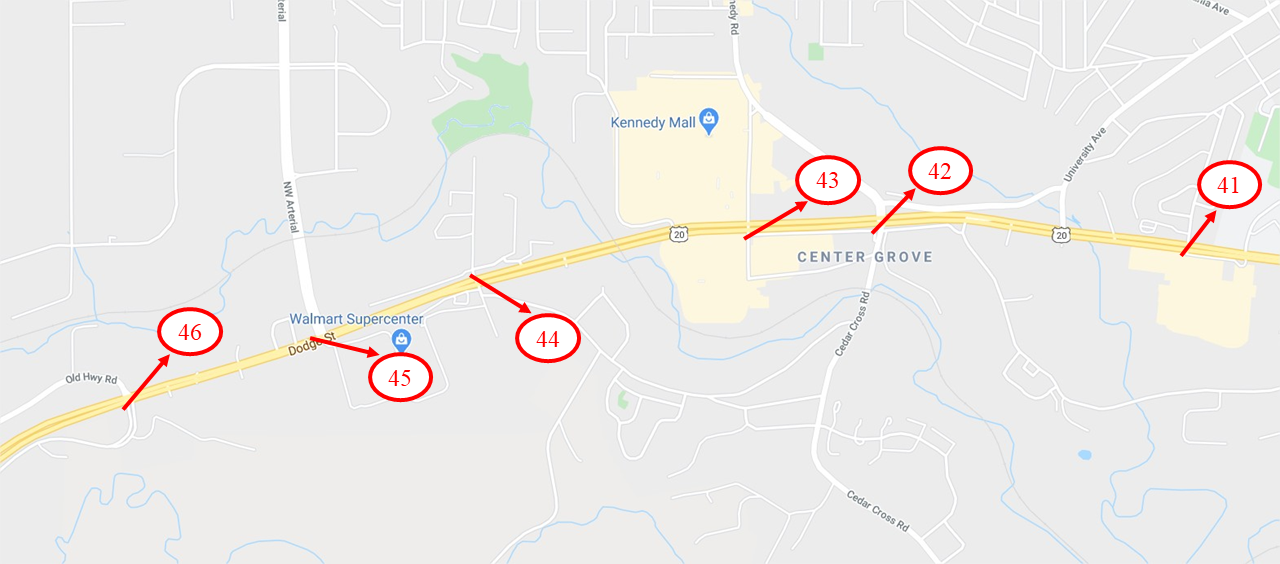}
        \caption{AI City Challenge camera map}
        \label{fig:cam_topology_aic}
     \end{subfigure}
     \\
     \begin{subfigure}{0.45\textwidth}
        \includegraphics[width=\textwidth,height=0.55\textwidth]{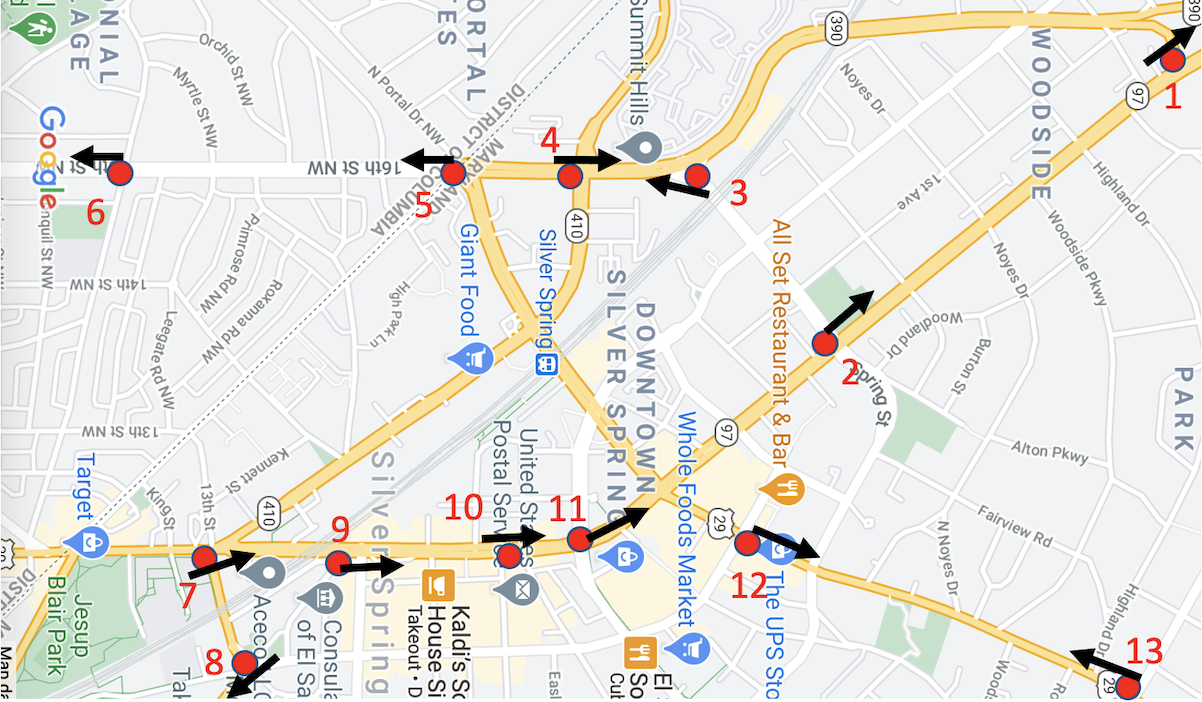}
        \caption{Sample RITIS system camera map}
     \end{subfigure}
     \caption{Traffic cameras topology can help us reduce search to adjacent cameras only.}
     \label{fig:cam_topology}
\end{figure}

To train EVER, we adopt both Cross Entropy and Triplet \cite{hermans2017defense} loss functions which is a common approach to train re-identification models. To make a training batch, we randomly sample $K$ vehicle tracks with unique identities out of all the training tracks and for each selected track $L$ frames are randomly selected and temporally ordered, \emph{i.e.} batch size is $K*L$. In addition, all the selected frames are resized to a fixed width $W$ and height $H$. As a result the computed batch is a tensor of shape $(K*L)*3*H*W$. Here we use the batch-hard sampling method for triplet loss formulated as below:
\begin{equation}
    \label{eq:triplet}
    \centering
    \small{
    \mathcal{L}_{t} = \frac{1}{B} \sum_{i=1}^{B} \sum_{a \in b_{i}} \left[\gamma + \max_{p \in \mathcal{P}(a)} {||f_a-f_p||}_2 - \min_{n \in \mathcal{N}(a)} {||f_a-f_n||}_2  \right]_{+}
    }
\end{equation}
In Eq. \ref{eq:triplet}, $B$, $b_i$, $a$, $\gamma$, $\mathcal{P}(a)$ and $\mathcal{N}(a)$ are the total number of batches, $i^{th}$ batch, anchor sample, distance margin threshold, positive and negative sample sets corresponding to a given anchor respectively. Moreover, $f_a, f_p, f_n$ are the extracted features for anchor, positive and negative samples. In addition, the Cross entropy loss with label smoothing technique \cite{szegedy2016rethinking} is used to alleviate the issue of over-fitting. Note that to effectively apply both cross entropy and triplet objectives to the extracted features, Batch Normalization Neck (BNNECK) \cite{Luo_2019_CVPR_Workshops} is employed. The Cross entropy loss is calculated as follows:
\begin{equation}
    \label{eq:cls}
    \centering
    \mathcal{L}_{c} = -\frac{1}{N} \sum_{i=1}^{N} \sum_{j=1}^{C} y^i_j\log\hat{y}^i_j 
    %+ (1 - y^i_j)\log(1 - \hat{y}^i_j)\right]
\end{equation}
Where $ \hat{y}^i_j = \frac{e^{(W_{j}^T f_i + b_{j})}}{\left(\sum_{k=1}^{C}e^{W_{k}^T f_i + b_k}\right)}$ is the  computed logit corresponding to class $j$ for the extracted feature $f_i$ of the $i^{th}$ training sample after applying the softmax layer. Furthermore, $W_{j}$, $b_j$ are the classifier's weight vector and bias associated with $j^{th}$ class respectively, and $N$ and $C$ represent the total number of samples and classes in the training dataset. Since we use label smoothing, $y^i_j = 1 - \frac{C-1}{C}\epsilon$ if $j=c$, otherwise $y^i_j = \frac{\epsilon}{C}$ where $c$ is the true label of $i^{th}$ sample and $\epsilon\in[0,1]$ is a hyper-parameter.

\textit{Camera Bias Mitigation:}
Another issue in multi-camera re-identification is that the orientation and background bias usually infiltrate to the computed vehicle embeddings from each camera  \cite{Zhu_2020_CVPR_Workshops} due to the limited variability. To alleviate this problem, once the model is trained, similar to \cite{Luo_2021_CVPR}, first we average all the vehicle representations $\{f^c_i\}^{N}_{i=1}$ captured in a particular camera $c$, resulting in a camera embedding $g_c = \frac{\sum_{i=1}^{N}f^c_i}{N}$. As many representations participate in this averaging, identity-dependent information can be suppressed while camera-dependent information is retained. Therefore, we can reduce the impact of the camera background and orientation bias on the re-identification and cross-camera comparison by subtracting a portion $\lambda$ of camera embedding $g_c$ from vehicle representations of that camera:
\begin{equation}
    f_i = f^c_i - \lambda g_c
\end{equation}

where $\lambda$ is a hyper-parameter. Finally, we employ the re-ranking method of \cite{zhong2017re}, a common post-processing technique to enhance the distance matrix and re-identification results concurrently. 

\subsection{Multi-Camera Tracker}
The multi-camera tracker is responsible for collecting the terminated single camera tracks from all the cameras and associate them into multi-camera tracks. To do so, this module has distinct sub-modules that are described below.

\subsubsection{Single Camera Track Collection} 
As each single camera is processed in parallel and concluded single camera tracks are obtained independently, a supervisory module is required to collect the concluded tracks from all the camera processes at designated time steps depending on the query frequency in the multi-camera tracking system. Afterwards, the collected single camera tracks and existing multi-camera tracks are passed on to the next stage to compute pairwise similarities.

\subsubsection{Track Similarity Matrix Calculation}
To create associations among the different vehicle tracks, we need to measure pairwise similarity of tracks from the perspectives of extracted deep features and spatio-temporal information. The re-identification module is trained and expected to extract embeddings to be similar using either Cosine or Euclidean distance for the same identities while embeddings corresponding to different identities to be dissimilar. However, due to the limited variability of vehicle types and models, lack of highly discriminating features and inherent camera view and orientation bias in embeddings, mere consideration of embedding similarity may result in erroneous multi-camera tracks. Therefore, spatio-temporal information is a valuable source of information needed to refine the similarity results. To this end, the following traffic rules are considered for the incorporation of spatio-temporal information and measuring the similarity of each pair of tracks:

\begin{figure}
    \centering
    \includegraphics[width=0.4\textwidth]{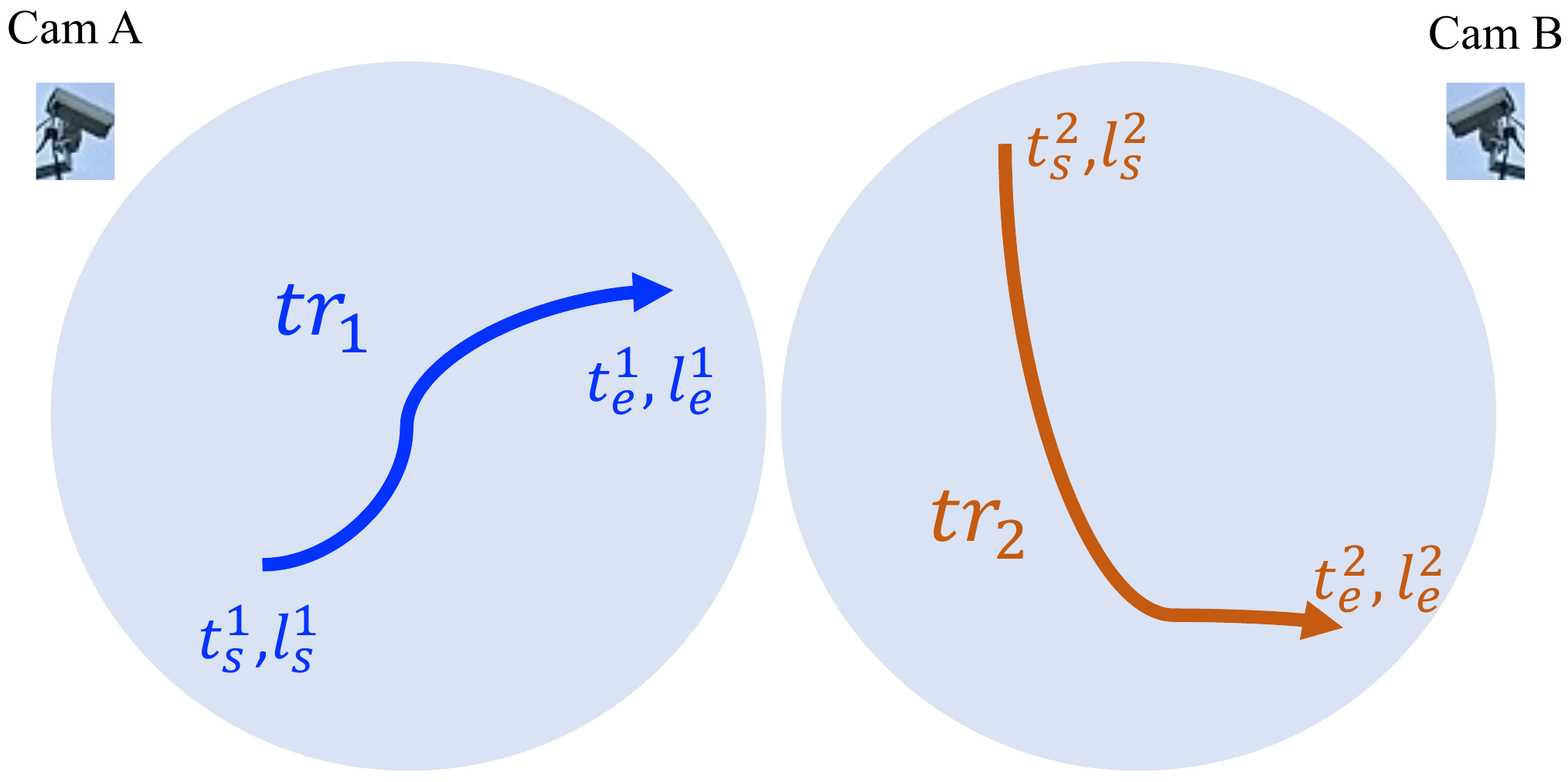}
    \caption{In case ${tr}_1$ and ${tr}_2$ with $t^{2}_s > t^{1}_e$ represent the same identity at neighboring cameras, the following relationships should be valid: $d(l^{1}_{s}, l^{2}_{s}) \geq d(l^{1}_{e}, l^{2}_{s})$, and $d(l^{2}_{e}, l^{1}_{e}) \geq d(l^{2}_{s}, l^{1}_{e})$. These criteria ensure that the direction of travel across cameras stay consistent going forward and backward in time.}
    \label{fig:tr_dir_consistency}
\end{figure}

\begin{itemize}
    \item [1-] Tracks captured through a same camera share similarities in orientation, shape and background that can negatively impact the re-identification results by severely reducing the inter-class distance which may lead to failure cases. Therefore, single camera tracks from the same camera cannot be matched together. Given the recent improvements in the area of Multi-object single camera tracking the chance of occurring ID switches has significantly reduced. Note that there is a chance that a passing vehicle might get back to the scene after a while. This can be fixed by setting a minimum time limit to consider single camera tracks from the same camera. However, as this adds additional hyper-parameters to the system, for the sake of simplicity we assume that the chance of reappearance is zero in this work. This is a reasonable assumption as the multi-camera tracks that are inactive for a while are flushed out of the system.
    \item [2-] In the event that the two tracks are captured by two non-overlapping cameras, their occurrence time should not overlap. Otherwise, their similarity should be set to zero.
    \item [3-] Vehicle's travel velocity should be within a reasonable range that can be determined for each particular scenario. In our implementation, similar to \cite{he2019multi} we considered track similarity to be a quadratic function of speed, \emph{i.e.} ${sim}_{v} = \max(0, 4\bar{v}(v_{max} - \bar{v})/v_{max})$ where $v_{max}$ and $\bar{v}=d(l^{2}_{s}, l^{1}_{e}) / (t^{2}_{s} - t^{1}_{e})$ are the maximum possible and the average travel speeds between the two locations where tracks ${tr}_1$ and ${tr}_2$ are observed. Start and end time-location of ${tr}_1$ are $(t^{1}_{s}, l^{1}_{s})$ and $(t^{1}_{e}, l^{1}_{e})$. Similarly, $(t^{2}_{s}, l^{2}_{s})$ and $(t^{2}_{e}, l^{2}_{e})$ represent the start and end time-location of ${tr}_2$. Here we assume that $t^{2}_s > t^{1}_e$. Also $d(.,.)$ represents the physical distance between the two points using their latitude and longitude. To obtain latitude and longitude of points in the image domain, extrinsic camera calibration can be performed using the Perspective-n-Point (PnP) technique and providing a set of corresponding points GPS and 2D pixel locations. This enables us to compute the homography matrix and its inverse to project points from real-world to 2D pixel coordinates and vice versa.
    \item [4-] Given the topology of traffic cameras, we can further reduce the search to only tracks that are observed in adjacent cameras. This constraint significantly reduces the number of comparisons needed for association of tracks while improving the accuracy. Figure \ref{fig:cam_topology} shows that the topology of cameras can be a valuable source of information which is readily available as latitude and longitude of traffic cameras are typically provided. 
    \item [5-] Since we reduce the search to only neighboring cameras in traffic rule 4 and assuming relatively straight roads, if two different tracks, ${tr}_1$ and ${tr}_2$ as shown in Figure \ref{fig:tr_dir_consistency}, represent the same vehicle identity, their travel direction should stay consistent in adjacent cameras both going forward and backward in time. Assuming $t^{2}_s > t^{1}_e$ and they represent the same vehicle identity, the following should hold true:
    \begin{equation}
        d(l^{1}_{s}, l^{2}_{s}) \geq d(l^{1}_{e}, l^{2}_{s}), \quad d(l^{2}_{e}, l^{1}_{e}) \geq d(l^{2}_{s}, l^{1}_{e})
    \end{equation}
\end{itemize}

Based on the above-mentioned traffic rules, the pairwise similarity of tracks ${tr}_{i}$ and ${tr}_j$ can be computed as follows:
\begin{equation}
    sim(i,j) = \begin{cases} 
      (1 - \frac{{||f_{i} - f_{j}||}_2}{2}) {sim}_{v} & \text{Traffic rules are satisfied} \\ 
      0 &  \text{otherwise}
   \end{cases}
\end{equation}
Note that the deep visual embeddings $f_{i}$ and $f_{j}$ are normalized \emph{i.e.} ${||f_i||}_2 = 1$ and ${||f_j||}_2 = 1$. This formulation allows us to compute the similarity matrix for all the pairs of tracks that are being considered for association. Note that, all the tracks are considered as both query and gallery tracks. Therefore, the resulting similarity matrix is symmetrical and all diagonal elements are set to zero. 
\begin{figure*}[h!]
    \begin{subfigure}[b]{0.24\textwidth}
    \centering \includegraphics[width=\textwidth]{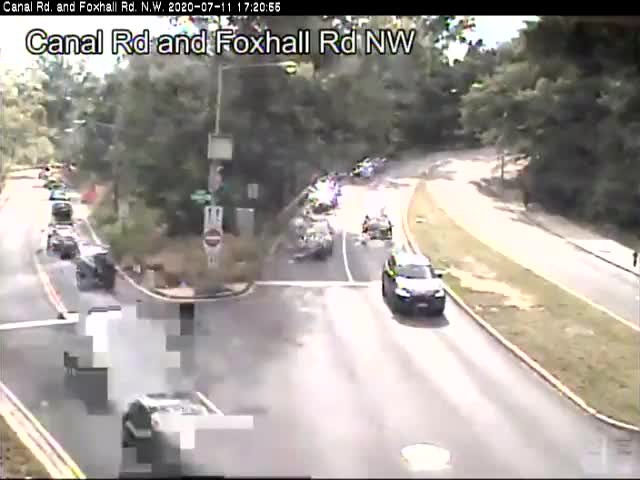}
    \caption{}
    \end{subfigure}
    \begin{subfigure}[b]{0.24\textwidth}
    \centering \includegraphics[width=\textwidth]{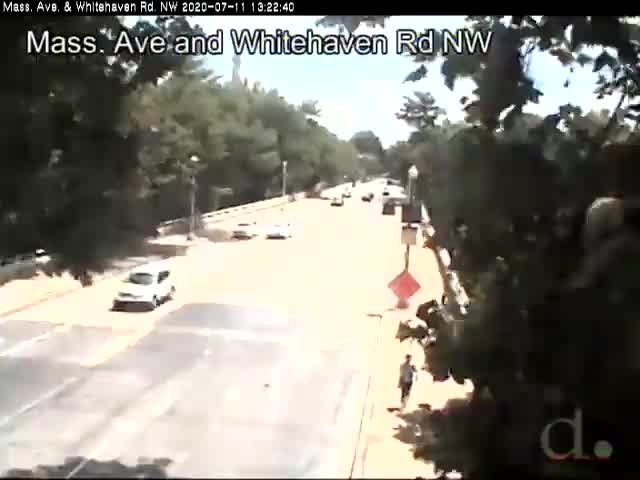}
    \caption{}
    \end{subfigure}
    \begin{subfigure}[b]{0.24\textwidth}
    \centering \includegraphics[width=\textwidth]{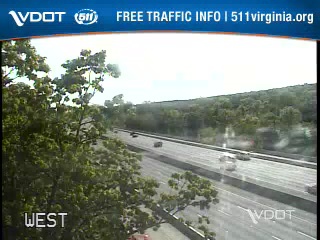}
    \caption{}
    \end{subfigure}
    \begin{subfigure}[b]{0.24\textwidth}
    \centering \includegraphics[width=\textwidth]{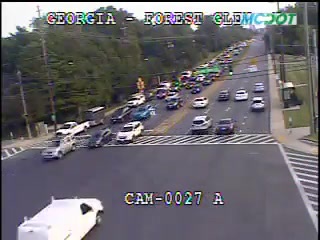}
    \caption{}
    \end{subfigure}
    \caption{Sample frames of traffic data obtained from RITIS platform. The low resolution, motion and compression artifacts are hallmarks of the streaming data on this platform.}
    \label{fig:camerastills}
\end{figure*}
\subsubsection{Hierarchical Clustering}
Since the number of true identities is not known beforehand, we adopt a hierarchical clustering algorithm to perform this task. Once the similarity matrix is computed we need to identify those tracks representing same vehicle identities. First, we enforce that a minimum similarity value to be met in order for tracks to represent identities. Therefore, depending on the dataset, we apply a minimum threshold $\tau_{min}$ to zero out similarity values smaller than $\tau_{min}$. Afterwards, we use hierarchical clustering algorithm which initiates each track as its own cluster and in an iterative manner merges the clusters with highest similarity. Note that the first traffic rule asserts that no more than one single camera track from the same camera can represent a vehicle identity. Therefore, once a track, \emph{e.g.} ${tr}_i$ from camera $i$, is merged with another track, \emph{e.g.} ${tr}_j$ from camera $j$, then we need to make sure that ${tr}_i$ cannot be merged with any other track from camera $j$ and track ${tr}_j$ cannot be merged with any other track from camera $i$. This operation helps us to maintain the transitivity property among all merged tracks.

\section{Experimental Results and Implementation Details}\label{sec:Experiments}
In this section we evaluate our proposed real-time multi-camera tracking system on the traffic data streaming on the RITIS platform as well as the 2021 AI City challenge multi-camera tracking dataset. Details of each type of data, the implementation details, justification for design choices, and discussions are provided accordingly. 

\subsection{RITIS Platform Streaming Data}
\subsubsection{Dataset} The RITIS system consolidates and rebroadcasts traffic data streams from various local municipalities throughout the United States on its platform. The streaming data provides typical metadata such as resolution, frame rate, and GPS coordinate of the camera. Sample video stills are shown in Figure \ref{fig:camerastills}. This data has hallmarks of true operational data including low resolution (usually $320\text{x}240$), low frame rate (usually $10$ fps or lower), and motion and compression artifacts. A visual inspection shows that the distribution of this data significantly differs from that of other curated traffic datasets. For optimal performance on detection, \emph{i.e.} the most fundamental task in our proposed multi-camera tracking pipeline, it is important to leverage all the available labeled data while accounting for the domain gap. Therefore, we employ labeled traffic datasets discussed in \ref{subsubsubsec:labled_data} to perform domain adaptation and train our vehicle detection model. 

\textit{Labeled Data}:\label{subsubsubsec:labled_data}
Two external labeled datasets are used for training the vehicle detector which will be referred to as our labeled source data.
\begin{itemize}
    \item UA-DETRAC \cite{CVIU_UA-DETRAC} dataset contains over ten hours of video from twenty-four different traffic cameras in Beijing and Tianjian, China. The videos were recorded at twenty-five frames per second, at the resolution of $960\text{x}540$, and under various lighting conditions. Of the collected data, $140$k frames were annotated which resulted in 1.21 million bounding boxes for $8250$ vehicles.
    \item CityCam \cite{citycam} dataset contains recordings from $212$ traffic cameras in the United States, with a resolution of $352\text{x}240$ and a frame rate of $1$ frame/per second. A total of $60$k frames and $900$k objects were annotated.
\end{itemize}

\textit{Unlabeled Data}:
Instead of operating on continuously streaming data, we collected a dataset which we refer to as our unlabeled target data. This dataset contains one-minute video recordings captured by 62 different cameras in the Washington D.C - Maryland - Virginia area. The recordings were captured multiple times during a day. Since there are no annotations available for this dataset, to evaluate our algorithms on target data, we curated a small detection validation set containing 1408 images from 62 cameras, with an approximately equal number of images per camera. From these images, 11,147 vehicles were labeled as ground truth. The details of labelling is provided in Table \ref{tab:catt_detection_statistics}. Additionally for the tracking task, nine 1-minutes videos were labeled in their entirety as our single camera tracking validation set; the summary of labeled boxes and tracks is presented in Table \ref{tab:tracker_statistics}. 

\begin{table}[ht]
    \caption{Detection validation set statistics}
    \label{tab:catt_detection_statistics}
    \centering
    \setlength\tabcolsep{12pt}
    \resizebox{\columnwidth}{!}{
    \begin{tabular}{c|c|c|c|c|c|c}
    \cline{2-7}
    \multicolumn{1}{c|}{} & Car & Bus & Truck & Van & SUV &  \multicolumn{1}{|c|}{\textbf{Total}} \\
    \cline{1-7}
    \multicolumn{1}{|c|}{Boxes} & $9659$ & $74$ & $283$ & $655$ & $476$ & \multicolumn{1}{|c|}{$11147$}\\
    \hline
    \end{tabular}
    }
\end{table}
\begin{table}[ht]
    \caption{Single camera tracking validation set statistics}
    \label{tab:tracker_statistics}
    \centering
    \setlength\tabcolsep{12pt}
    \resizebox{\columnwidth}{!}{
    \begin{tabular}{c|c|c|c|c|c|c}
    \cline{2-7}
    \multicolumn{1}{c|}{} & Car & Bus & Truck & Van & SUV &  \multicolumn{1}{|c|}{\textbf{Total}} \\
    \cline{1-7}
    \multicolumn{1}{|c|}{Boxes} & $22296$ & $555$ & $1487$ & $2375$ & $518$ & \multicolumn{1}{|c|}{$27231$}\\
    \cline{1-7}
    \multicolumn{1}{|c|}{Tracks} & $157$ & $3$ & $11$ & $11$ & $5$ & \multicolumn{1}{|c|}{$187$} \\
    \hline
    \end{tabular}
    }
\end{table}

\subsubsection{Multi-Vehicle Detection} For this data, we adopt RetinaNet \cite{lin2018focal} and Faster R-CNN \cite{fasterRcnn} trained on the COCO dataset. As pointed out before, the characteristics of RITIS streaming data are quite different compared to other publicly available benchmark datasets. Therefore, there is a domain shift which can hinder the performance. Moreover, there are no annotations available for this dataset. However, annotations such as bounding box information, transfer across domains. Hence unsupervised domain adaptation from our labeled data to unlabeled data, can help us to compensate for this gap and prepare training data for the vehicle detection module as discussed in the following section.
\begin{figure*}[h!]
\centering
\begin{subfigure}{0.24\linewidth}
  \centering
  \includegraphics[width=\textwidth]{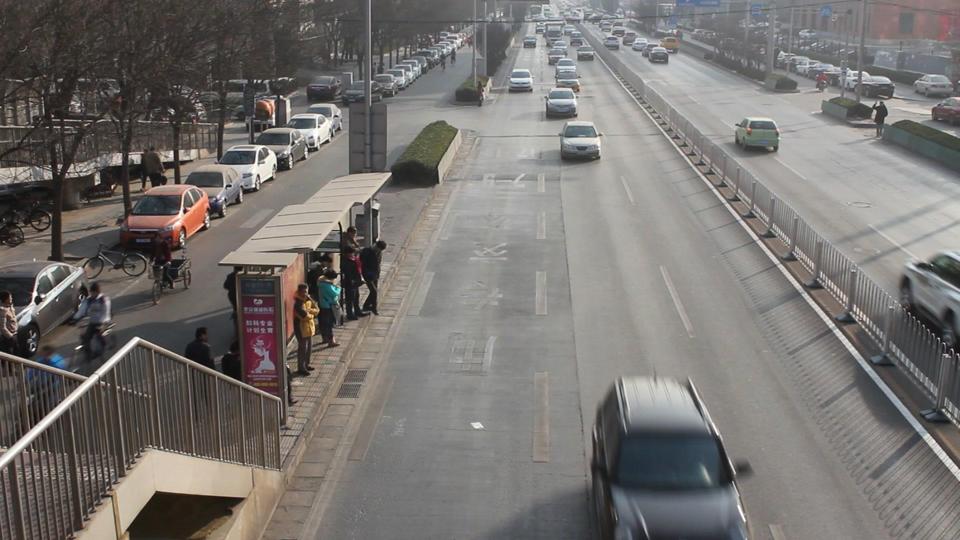}
  \caption{Original image}\label{fig:originaluadetrac}
\end{subfigure} 
\begin{subfigure}{.24\linewidth}
    \centering
    \includegraphics[width=\textwidth]{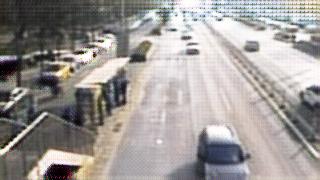}
    \caption{DA after 10k iterations}\label{fig:dauadetrac10k}
\end{subfigure}
\begin{subfigure}{.24\linewidth}
    \centering
    \includegraphics[width=\textwidth]{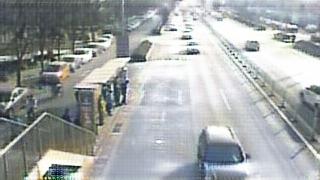}
    \caption{DA after 50k iterations}\label{fig:dauadetrac50k}
\end{subfigure}
\begin{subfigure}{.24\linewidth}
    \centering
    \includegraphics[width=\textwidth]{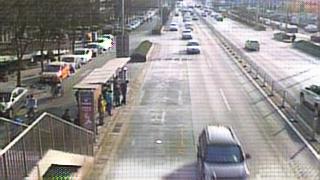}
    \caption{DA after 100k iterations}\label{fig:dauadetrac100k}
\end{subfigure}
\caption{The domain transfer applied to the UA-DETRAC. (a) shows the original image, while figures (b)-(d) shows the extent of domain adaptation during the course of training.}
\label{fig:daexamples}
\end{figure*}
\begin{table}[]
    \caption{The results of training the detector. The data is fine-tuned on the specified dataset and tested on the operational traffic camera data. ``All" refers to training on the combination of the best performing domain-adapted UA-DETRAC dataset and the CityCam Dataset.}
    \label{tab:detectonall}
    \resizebox{1.\columnwidth}{!}{
    \begin{tabular}{c|c|c|c|c|c}
         \cline{2-6} & \multicolumn{5}{|c|}{Dataset} \\
         \hline
         \multicolumn{1}{|c|}{\multirow{2}{*}{Model}} &
         \multicolumn{1}{|c|}{Baseline} &
         \multicolumn{1}{|c|}{CityCam} &
         \multicolumn{1}{|c|}{UA-DETRAC} &
          \multicolumn{1}{|c|}{DA\_UA-DETRAC} &
         \multicolumn{1}{|c|}{All}\\
        \cline{2-6}
         \multicolumn{1}{|c|}{} &
         \multicolumn{1}{|c|}{mAP(\%)} &
         \multicolumn{1}{|c|}{mAP(\%)} &
         \multicolumn{1}{|c|}{mAP(\%)} &
         \multicolumn{1}{|c|}{mAP(\%)} &
         \multicolumn{1}{|c|}{mAP(\%)} \\
        \cline{1-6}
        \multicolumn{1}{|c|}{FasterRCNN-101} &
        \multicolumn{1}{|c|}{$64.8$} &
        \multicolumn{1}{|c|}{$73.92$} &
        \multicolumn{1}{|c|}{$37.86$} &
        \multicolumn{1}{|c|}{$61.49$} &
        \multicolumn{1}{|c|}{$\textbf{77.83}$}\\
        \cline{1-6}
        \multicolumn{1}{|c|}{RetinaNet-101} &
        \multicolumn{1}{|c|}{$62.24$} &
        \multicolumn{1}{|c|}{$69.95$} &
        \multicolumn{1}{|c|}{$36.21$} &
        \multicolumn{1}{|c|}{$60.62$} &
        \multicolumn{1}{|c|}{$75.84$} \\
        \hline
    \end{tabular}
    }
\end{table}

\begin{figure}
    \centering 
    \includegraphics[width=0.4\textwidth]{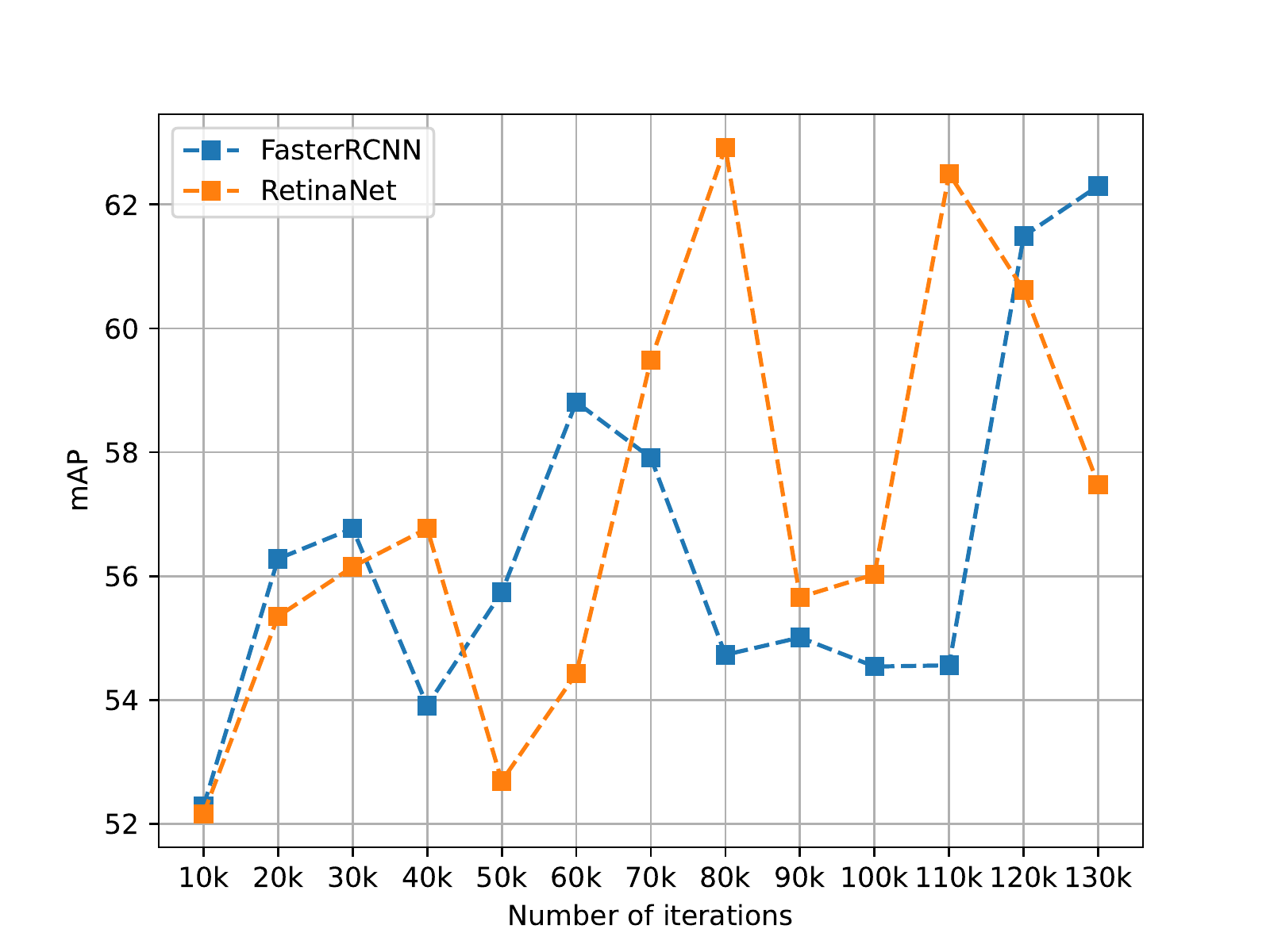}
    \caption{Detection performance vs number of iterations of domain adaptation.}\label{fig:da}
\end{figure}

\textit{Domain Adaptation:} \label{subsubsubsec:domainadaptation}
A typical procedure to leverage labeled source data and a pre-trained detector is to fine-tune on the labeled source data directly and test on the target data. However, domain shift can deteriorate performance. For instance, as shown in Table \ref{tab:detectonall}, when off-the-shelf RetinaNet and Faster R-CNN models are applied to our validation set as baseline models, the detection mAP is $62.24\%$ and $64.8\%$ respectively. However, once these detectors are fine-tuned on UA-DETRAC dataset, the performance drops to $36.21\%$ and $37.86\%$ correspondingly, showing the gap between the two domains. To leverage all available data, we use CycleGAN \cite{zhu2020unpaired} to perform unpaired image-to-image translation to transfer domain information from the target data into the UA-DETRAC labeled source data while maintaining the inherent structure and labels of the source data. Note that no domain adaptation is needed for the CityCam dataset as fine-tuning directly improved detection results on the held out validation set over the baseline as can be seen in Table \ref{tab:detectonall}. After we learn the mapping, we transfer the UA-DETRAC dataset to the domain of target data and reference it as DA\_UA-DETRAC. Figure \ref{fig:daexamples} demonstrates a sample image of UA-DETRAC dataset and its progression during the course of domain adaptation process. To measure the success of domain adaptation and control how much style was transferred, we use the detection accuracy as a proxy task. To do so, after every $10K^{th}$ iteration we record the checkpoint for the mapping function, train the object detector on that particular mapped UA-DETRAC dataset, and calculate the accuracy of detection. The mAP scores after each $10k^{th}$ iteration are shown in Figure \ref{fig:da}. We then choose to use the domain-adapted datasets corresponding to the highest performance. Table \ref{tab:detectonall} shows that domain adaptation leads to a nearly 24 point increase in detection accuracy once switched from UA-DETRAC to DA\_UA-DETRAC.

After we obtain DA\_UA-DETRAC, we combine this with the Citycam dataset to increase the size of the training set and get the best performance. Next we fine-tune both RetinaNet101 and Faster R-CNN101 models on this combined data. Note that to train object detectors, both the stem and the first residual stage of the networks are frozen while the rest of the networks are trained. Stochastic Gradient Descent optimizer with learning rate of $0.0025$ and momentum of $0.9$ is used for $20K$ iterations. Table \ref{tab:detectonall} reports that domain adaptation of DA\_UA-DETRAC along with combination with the CityCam dataset help to increase the detection accuracy approximately $13\%$ from the baseline models. RetinaNet101 and Faster R-CNN101, with a single NVIDIA RTX 2080 Ti GPU card, need $68$ and $69$ milliseconds on average to process frames from RITIS data respectively. As Faster R-CNN101 achieved higher mAP on our validation set, we choose it to be our detection model in the multi-camera vehicle tracking. This helps us to meet the real-time requirements of $100$ milliseconds ($10$ fps). 

%Given two discriminator networks $D_X, D_Y$ and mappings $G:X\rightarrow Y$ and $F:Y\rightarrowX$, we aim to minimize the forward consistency loss $x\rightarrow G(x) \rightarrow F(G(x)) \approx x$ and backward consistency loss $y \rightarrow F(y) \rightarrow  G(F(y)) \approx y$ such that origin of outputs from the generating functions can not be distinguished by the discriminators. 

%\begin{figure}
%    \centering
%    \includegraphics[width=0.48\textwidth]{graphs/bar_graphs.png}
%    \caption{Performance of trained models. ``All" refers to training on DA-UADETRAC and %CityCam}\label{fig:bargraphs}
%\end{figure}
            
\subsubsection{Single Camera Tracking}\label{sec:methodSCT}
%Assigning consistent identities from the detections is a critical step for multi-camera tracking. These detections, along with their corresponding deep features from a ResNet101\_IBN-a backbone, were associated temporally using DeepSORT \cite{wojke2017simple}. This module takes state information such as bounding box centers, aspect ratios, and height (along with their velocities) as input to a Kalman filter; the predictions from the Kalman filter, along with the accumulated deep features, is then matched to tracks according to the Hungarian algorithm. This model assumes the bounding boxes change at a constant velocity; while this is not always satisfied, the matching of deep features provides some robustness against errors.
Here we use standard evaluation metrics, \emph{i.e.} Multiple Object Tracking Accuracy (MOTA) and ${IDF}_1$ scores, as described in \cite{motMetrics} to evaluate the single camera tracking performance. MOTA is a function of the number of false negatives, false positives, fragmentations, and true detections. MOTA accounts for the frequency of mismatches; however, it is important to consider how long these errors occur, which is enumerated by the ${IDF}_1$ score \cite{motMetrics}. As discussed in section \ref{subsubsec:sct_method}, we use a modified version of DeepSort to associate detected objects in a single camera. We use our tracking validation set with statistics presented in Table \ref{tab:tracker_statistics} to determine the best set of hyper-parameters required by this tracker. We use the Euclidean distance measure for similarity and set our matching threshold to $0.3$, below which tracks are candidates for matching. %We begin a track after three detections are associated, and we store a track until fifteen consecutive frames pass without a track association. 
The results, as well as the desired direction of scores for each metric, are shown in table \ref{tab:cattTrackerStats}.
%The number of Ground Truth (GT) are the number of labeled tracks in each video,  while Mostly Tracked (MT) and Mostly Lost (ML) correspond to whether the tracks covered the ground truth at least 80\% or at most 20\%, respectively. False Positives (FP) occur when an object is considered tracked when no such object exists, and False Negatives (FN) occur when an object is present but the tracker does not track it. Identity Switching (IDs) measure how often the tracked objects' ID changes. 
\begin{table}[h!]
    \centering
    \caption{Single Camera Tracking results on our held out validation set}
    \label{tab:cattTrackerStats}
    \begin{tabular}{c|c|c|c|c}
    \cline{2-5}
    \multicolumn{1}{c|}{} & IDP($\uparrow$) & IDR($\uparrow$) & IDF1($\uparrow$) & \multicolumn{1}{c|}{MOTA ($\uparrow$)} \\
    \cline{1-5}
    \multicolumn{1}{|c|}{\textbf{Overall}} & $85.5$ & $74.5$ & $79.7$ & \multicolumn{1}{|c|}{$68.7$}  \\
    \hline
    \end{tabular}
\end{table}

\subsubsection{Deep Feature Extraction}
For the purpose of extracting frame-level deep features for the RITIS data, we use ResNet50\_IBN-a as the backbone architecture of EVER model. In contrast to the vehicle detection task which mainly requires large-scale information to localize vehicles, vehicle re-identification has to deal with small-scale and subtle details of vehicles. Therefore, employing domain adaptation for publicly available datasets to be transferred to the RITIS data might degrade microscopic features and negatively impact the re-identification process. Based on this, we decided to create a vehicle re-identification dataset directly from RITIS data. In total we collected $75,685$ images ($858$ tracks) of $475$ vehicle identities across $13$ cameras. $400$ of these vehicle identities which form $721$ vehicle tracks have been assigned for training and the rest and reserved for validation. Afterwards, we train the video-based version of EVER model as discussed in section \ref{subsubsec:vereid_method} and by the following hyper-parameters. We set the track temporal length to $L=5$. Also, the number of unique identities within each mini batch is set to $K=16$. Since the resolution of videos in this domain is relatively low, \emph{e.g.} $320\times240$, vehicles incorporate fewer pixels. Therefore, we resized vehicle crops to height and width $H=64$, and $W=64$ as opposed to $224\times224$ or $256\times256$ which are typically used. This in turn contributes to reducing the inference time. In fact, this model can compute frame-level deep features of 128 vehicle images in only $6.51$ milliseconds on a NVIDIA RTX 2080 Ti GPU card. Consequently, we perform horizontal-flip test-time augmentation technique, in which we also compute the representation of the horizontally flipped tensor of a track and average it with its original representation to obtain a more robust embedding. Table \ref{tab:ritis_reid_val} presents the track-based vehicle re-identification results on the validation set. Based on the observed performance, it can be seen that conducting re-identification in low resolution images with motion and compression artifacts is challenging. 
\begin{table}[]
    \centering
    \caption{Evaluation results of the trained EVER model on our RITIS validation set for the task of track-based vehicle re-identification.}
    \label{tab:ritis_reid_val}
    \begin{tabular}{c|c|c|c}
        \cline{1-4}
        \multicolumn{1}{|c|}{\multirow{2}{*}{Model Settings}} & \multicolumn{1}{|c|}{\multirow{2}{*}{mAP(\%)($\uparrow$)}} & \multicolumn{2}{|c|}{CMC(\%)($\uparrow$)} \\
        \cline{3-4}
        \multicolumn{1}{|c|}{} & \multicolumn{1}{|c|}{} & \multicolumn{1}{|c|}{@1} & \multicolumn{1}{|c|}{@5} \\
        \cline{1-4}
        \multicolumn{1}{|c|}{EVER-trained ResNet50\_IBN-a} & \multicolumn{1}{|c|}{$35.0$} & \multicolumn{1}{|c|}{$32.7$} & \multicolumn{1}{|c|}{$44.0$} \\
        \cline{1-4}
        \multicolumn{1}{|c|}{+ Horizontal Flip Augmentation} & \multicolumn{1}{|c|}{$37.4$} & \multicolumn{1}{|c|}{$33.5$} & \multicolumn{1}{|c|}{$44.2$} \\
        \cline{1-4}
        \multicolumn{1}{|c|}{+ Re-ranking} & \multicolumn{1}{|c|}{$38.1$} & \multicolumn{1}{|c|}{$34.0$} & \multicolumn{1}{|c|}{$44.1$} \\
        \cline{1-4}
        \multicolumn{1}{|c|}{+ Camera Bias Mitigation} & \multicolumn{1}{|c|}{$\textbf{38.5}$} & \multicolumn{1}{|c|}{$\textbf{34.6}$} & \multicolumn{1}{|c|}{$\textbf{44.3}$} \\
        \cline{1-4}
    \end{tabular}
\end{table}
Table \ref{tab:ritis_reid_val} shows the impactful roles of test-time horizontal flip augmentation and camera bias mitigation. 

\subsubsection{Multi-Camera Tracking} 
\begin{figure}
    \centering
    \includegraphics[width=0.49\textwidth]{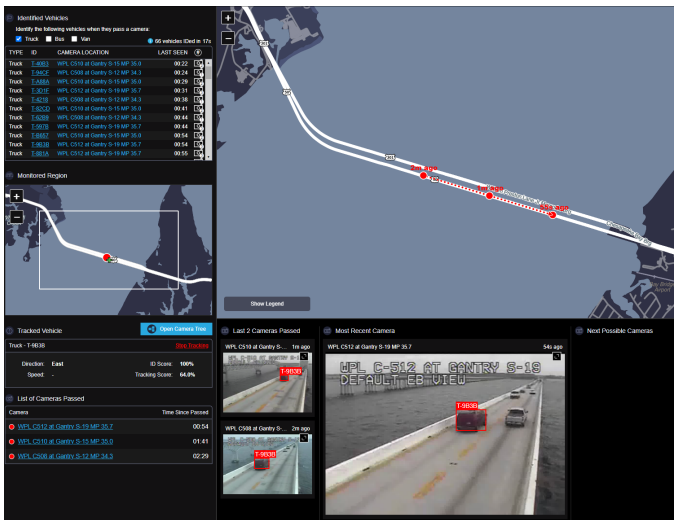}
    \caption{Screen shot of real-time multi-camera vehicle tracking capability on three sampled neighboring cameras on RITIS platform.}
    \label{fig:RITIS_prototype}
\end{figure}

Once all the required modules in our proposed multi-camera vehicle tracking system are prepared, we start tracking vehicle identities in the RITIS platform over the set of cameras that are of interest. Since RITIS gathers real-time traffic data streams, all the multi-camera processing needs to be completed in real-time to maintain time parity with streaming data and enable the end-user to make real-time decisions based on what is happening at the moment as waiting for a few seconds, minutes, and hours might not be an option. Our proposed method has been implemented on this platform as a prototype for multi-camera tracking and it is shown to be able to track vehicle identities in real-time. Figure \ref{fig:RITIS_prototype} shows a screen shot of the interface for this prototype. The highlighted track has been re-identified in the three cameras. 

\subsection{AI City Challenge}
\subsubsection{Dataset}
This multi-camera vehicle tracking dataset, namely CityFlow, contains $3.5$ hours of traffic videos collected from 46 cameras spanning $16$ intersections in a mid-sized city in Iowa, United States. The videos resolution are at least $960$p with $10$ frames per second. In addition, an online evaluation server is provided to rank participating teams based on the ${IDF}_1$ tracking metric which is the ratio of correctly identified tracks over the average number of ground-truth and computed tracks. In another words, ${IDF}_1$ balances identification precision and recall via computing their harmonic mean.

\subsubsection{Multi-Vehicle Detection}
For this dataset, we choose to use off-the-shelf object detector trained on COCO dataset as this data does not contain properties such as video tear or compression artifacts. RetinaNet, Faster R-CNN, Mask R-CNN and EfficientDet object detectors are considered. Table \ref{tab:det_aic_benchmark} presents the average inference time (On a single NVIDIA RTX 2080 GPU card using batch size of $1$) of these detectors on the high resolution AI City Challenge dataset along with their respective mean Average Precision (mAP) on the COCO 2017 benchmark. 
\begin{table}[]
    \caption{Inference time-Detection Accuracy comparison of popular object detection models. mAP is measured on the COCO 2017 benchmark. Inference time, represents the average time it takes for a single NVIDIA RTX 2080 Ti GPU card to process high resolution video frames of 2021 AI City Challenge with batch size of $1$.}
    \label{tab:det_aic_benchmark}
    \resizebox{\columnwidth}{!}{
    \begin{tabular}{c|c|c|c}
    \cline{1-4}
    \multicolumn{1}{|c|}{\multirow{2}{*}{model}} & \multicolumn{1}{|c|}{\multirow{2}{*}{Backbone}} & \multicolumn{1}{|c|}{\multirow{2}{*}{mAP(\%)($\uparrow$)}} & \multicolumn{1}{|c|}{Inference}  \\
    \multicolumn{1}{|c|}{} & \multicolumn{1}{|c|}{} & \multicolumn{1}{|c|}{} & \multicolumn{1}{|c|}{time(ms)($\downarrow$)} \\
    \cline{1-4}
    \multicolumn{1}{|c|}{RetinaNet} & \multicolumn{1}{|c|}{ResNet101} & \multicolumn{1}{|c|}{$40.4$} & \multicolumn{1}{|c|}{$66$} \\
    \cline{1-4}
    \multicolumn{1}{|c|}{\multirow{2}{*}{Faster R-CNN}} & \multicolumn{1}{|c|}{ResNet101} & \multicolumn{1}{|c|}{$42.0$} & \multicolumn{1}{|c|}{$68$} \\
    \cline{2-4}
    \multicolumn{1}{|c|}{} & \multicolumn{1}{|c|}{ResNext101} & \multicolumn{1}{|c|}{$43.0$} & \multicolumn{1}{|c|}{$110$} \\
    \cline{1-4}
    \multicolumn{1}{|c|}{\multirow{2}{*}{Mask R-CNN}} & \multicolumn{1}{|c|}{ResNet101} & \multicolumn{1}{|c|}{$42.9$} & \multicolumn{1}{|c|}{$69$} \\
    \cline{2-4}
    \multicolumn{1}{|c|}{} & \multicolumn{1}{|c|}{ResNext101} & \multicolumn{1}{|c|}{$44.3$} & \multicolumn{1}{|c|}{$113$} \\
    \cline{1-4}
    \multicolumn{1}{|c|}{\multirow{3}{*}{EfficientDet}} & \multicolumn{1}{|c|}{D3} & \multicolumn{1}{|c|}{$47.2$} & \multicolumn{1}{|c|}{$\textbf{41}$} \\
    \cline{2-4}
    \multicolumn{1}{|c|}{} & \multicolumn{1}{|c|}{D4} & \multicolumn{1}{|c|}{$49.7$} & \multicolumn{1}{|c|}{$66$} \\
    \cline{2-4}
    \multicolumn{1}{|c|}{} & \multicolumn{1}{|c|}{D5} & \multicolumn{1}{|c|}{$\textbf{51.5}$} & \multicolumn{1}{|c|}{$104$} \\
    \hline
    \end{tabular}
    }
\end{table}
Among these detectors, EfficientDet D3 is a viable choice as it has the least inference time, \emph{i.e.} suitable for real-time multi-camera tracking, while beating the detection accuracy of RetinaNet, Fatser R-CNN and Mask R-CNN. In addition, we set the object detector to only retain detections of only "car", "bus", and "truck" types with a minimum of $0.35$ detection confidence. NMS with IOU threshold of $0.85$ is applied to the detection results. 

\subsubsection{Deep Feature Extraction}
To extract the deep visual features for this dataset, we use the ResNet101\_IBN-a as the backbone architecture of the EVER model. To train this model, we use CityFlow-ReID and VehicleX \cite{Yao20VehicleX} that are provided for the multi-camera vehicle re-identification task. CityFlow-ReID contains $85,058$ images from $880$ vehicle identities. Out of these images, $52,717$ ($2173$ tracks) from $440$ vehicle identities form the training set and the rest are reserved for testing. We randomly sample $100$ identities from the training set as the validation set which gives rise to $11349$ images ($469$ tracks). VehicleX is a synthetic vehicle re-identification dataset which has over $190,000$ images ($14663$ tracks) of $1300$ vehicle identities which is only provided for training purposes. However as discussed in \cite{Luo_2021_CVPR,Khorramshahi_2021_CVPR}, there is a significant domain shift between this synthetic data and the real data used for evaluation. Therefore, similar to our approach for vehicle detection on the RITIS data, we used CycleGAN to reduce this domain gap and learn a generator function to translate synthetic images to real data domain. After this step, we gather all the real and domain adapted synthetic data in our training set to train EVER model and extract discriminative deep representations of a vehicle track as described in \ref{subsubsec:vereid_method} section. Note that for this dataset, we set the number of distinct identities in each batch $K=16$, the track length $L=5$, and height and width of images to $H=256, W=256$. After training, this model computes frame-level deep visual features to be used by single camera tracker as well as the track level representations to be used by the multi-camera tracker. Note that during inference, the forward pass of the ResNet101\_IBN-a for the batch size of $128$ takes $13.09$ milliseconds on average. This allows us to meet the real-time requirement of $100$ milliseconds ($10$ fps) for traffic videos while performing horizontal flip test-time augmentation technique. Table \ref{tab:aic_reid_val} reports the track-based vehicle re-identification performance on our validation set. It is shown that a model ensemble can significantly contribute to the performance \cite{Luo_2021_CVPR}. However, having an ensemble is a significant overhead that is computationally prohibitive for the real-time applications. Here we merely want to show the performance boost by considering 3 different models namely, ResNet101\_IBN-a trained on real and domain adapted synthetic data, ResNet101\_IBN-a only trained on real data, and a ResNet101 train on real and domain adapted synthetic data. While this improves the performance,it requires 3 times of the regular inference time or computational resources that is not possible in a real-time large-scale multi-camera tracking system unless expensive computational units are available. In addition, we note that camera bias mitigation technique increased CMC@1 metric by $1.1\%$. Improving CMC@1 is directly tied with the performance of MCT since in hierarchical clustering performed by the multi-camera tracker, in each iteration the pair with highest matching score, \emph{i.e.}, the first match, is the basis for association. 

\begin{table}[]
    \centering
    \caption{Evaluation result of the trained EVER model on our validation set for the task of track-based vehicle re-identification.}
    \label{tab:aic_reid_val}
    \begin{tabular}{c|c|c|c}
        \cline{1-4}
        \multicolumn{1}{|c|}{\multirow{2}{*}{Model Settings}} & \multicolumn{1}{|c|}{\multirow{2}{*}{mAP(\%)($\uparrow$)}} & \multicolumn{2}{|c|}{CMC(\%)($\uparrow$)} \\
        \cline{3-4}
        \multicolumn{1}{|c|}{} & \multicolumn{1}{|c|}{} & \multicolumn{1}{|c|}{@1} & \multicolumn{1}{|c|}{@5} \\
        \cline{1-4}
        \multicolumn{1}{|c|}{EVER-trained ResNet101\_IBN-a} & \multicolumn{1}{|c|}{$54.2$} & \multicolumn{1}{|c|}{$61.8$} & \multicolumn{1}{|c|}{$80.5$} \\
        \cline{1-4}
        \multicolumn{1}{|c|}{+ Horizontal Flip Augmentation} & \multicolumn{1}{|c|}{$54.6$} & \multicolumn{1}{|c|}{$62.2$} & \multicolumn{1}{|c|}{$80.7$} \\
        \cline{1-4}
        \multicolumn{1}{|c|}{+ Re-ranking} & \multicolumn{1}{|c|}{$56.6$} & \multicolumn{1}{|c|}{$62.9$} & \multicolumn{1}{|c|}{$81.7$} \\
        \cline{1-4}
        \multicolumn{1}{|c|}{+ Camera Bias Mitigation} & \multicolumn{1}{|c|}{$56.8$} & \multicolumn{1}{|c|}{$64.0$} & \multicolumn{1}{|c|}{$81.3$} \\
        \hline
        \hline
        \multicolumn{1}{|c|}{+ Model Ensemble} & \multicolumn{1}{|c|}{$\textbf{63.5}$} & \multicolumn{1}{|c|}{$\textbf{69.4}$} & \multicolumn{1}{|c|}{$\textbf{85.0}$} \\
        \cline{1-4}
    \end{tabular}
\end{table}

\subsubsection{Multi-Camera Tracking}
After putting all the modules in our proposed real-time multi-camera tracking system, we evaluate our approach on the test set of 2021 NVIDIA AI City Multi-camera Tracking challenge and compare it against submissions to the public leaderboard. Note that submissions are ranked for this challenge only based on ${IDF}_1$ tracking score without any consideration for processing time and computational complexity. In addition, the evaluation server provides Identification Precision (IDP) and Identification Recall (IDR) evaluation metrics. The test set is composed of $6$ simultaneously recorded videos of resolution $1280\text{x}960$ from $6$ different cameras as shown in Figure \ref{fig:cam_topology_aic}. The length of each video is $3$ minutes and $20$ seconds. Table \ref{tab:multi_cam_track} summarizes our performance on this dataset and shows the impact of the traffic rules we considered during similarity matrix computation. The baseline model refers to our proposed system without the consideration of traffic rules $\#4$ \& $\#5$.
\begin{table}[]
    \centering
    \caption{Our proposed real-time multi-camera tracking results on the test set of 2021 AI City Challenge.}
    \label{tab:multi_cam_track}
    \begin{tabular}{c|c|c|c}
    \cline{1-4}
    \multicolumn{1}{|c|}{Settings} & $IDP$($\uparrow$) & $IDR$($\uparrow$) & \multicolumn{1}{|c|}{${IDF}_1$($\uparrow$)} 
    \\
       \cline{1-4}
       \multicolumn{1}{|c|}{Baseline} & $0.5207$ & $0.4469$ & \multicolumn{1}{|c|}{$0.4810$} \\
       \cline{1-4}
       \multicolumn{1}{|c|}{+ Traffic Rule \#4} & $0.5572$ & $0.6196$ & \multicolumn{1}{|c|}{$0.5867$} \\
       \cline{1-4}
       \multicolumn{1}{|c|}{+ Traffic Rule \#5} & $0.6608$ & $0.6890$ & \multicolumn{1}{|c|}{$0.6746$} \\
       \cline{1-4}
       \multicolumn{1}{|c|}{+ Hyper-parameter Tuning} & $\textbf{0.7088}$ & $\textbf{0.7292}$ & \multicolumn{1}{|c|}{$\textbf{0.7189}$} \\
       \hline
    \end{tabular}
\end{table}
From Table \ref{tab:multi_cam_track} we can appreciate the impact of effective incorporation of spatio-temporal information on the refinement of multi-camera tracking results. Most notably, by reducing the search only to neighboring cameras, \emph{i.e.} Traffic Rule $\#4$, there is a significant boost in the IDR metric showing that capability of recalling the multi-camera tracks is very well improved. Moreover, It can be seen that  enforcing the consistency in the travel direction (Traffic Rule $\#5$) is a valuable spatio-temporal cue to filter out erroneous results. Finally, in the public leaderboard of the challenge our method with the ${IDF}_1$ score of $0.7189$ is ranked $5^{th}$ out of 23 participating teams without any considerations for the computational efficiency, inference time, and entering additional spatio-temporal information beyond what is provided by the challenge organizers. For instance, \cite{Liu_2021_CVPR} shows that introducing crossroad zones, \emph{i.e.} the connectivity of the different road zones in one camera to the neighboring cameras, search for multi-camera tracks can be extremely refined.
\begin{figure}
    \centering
    \includegraphics[width=0.5\textwidth]{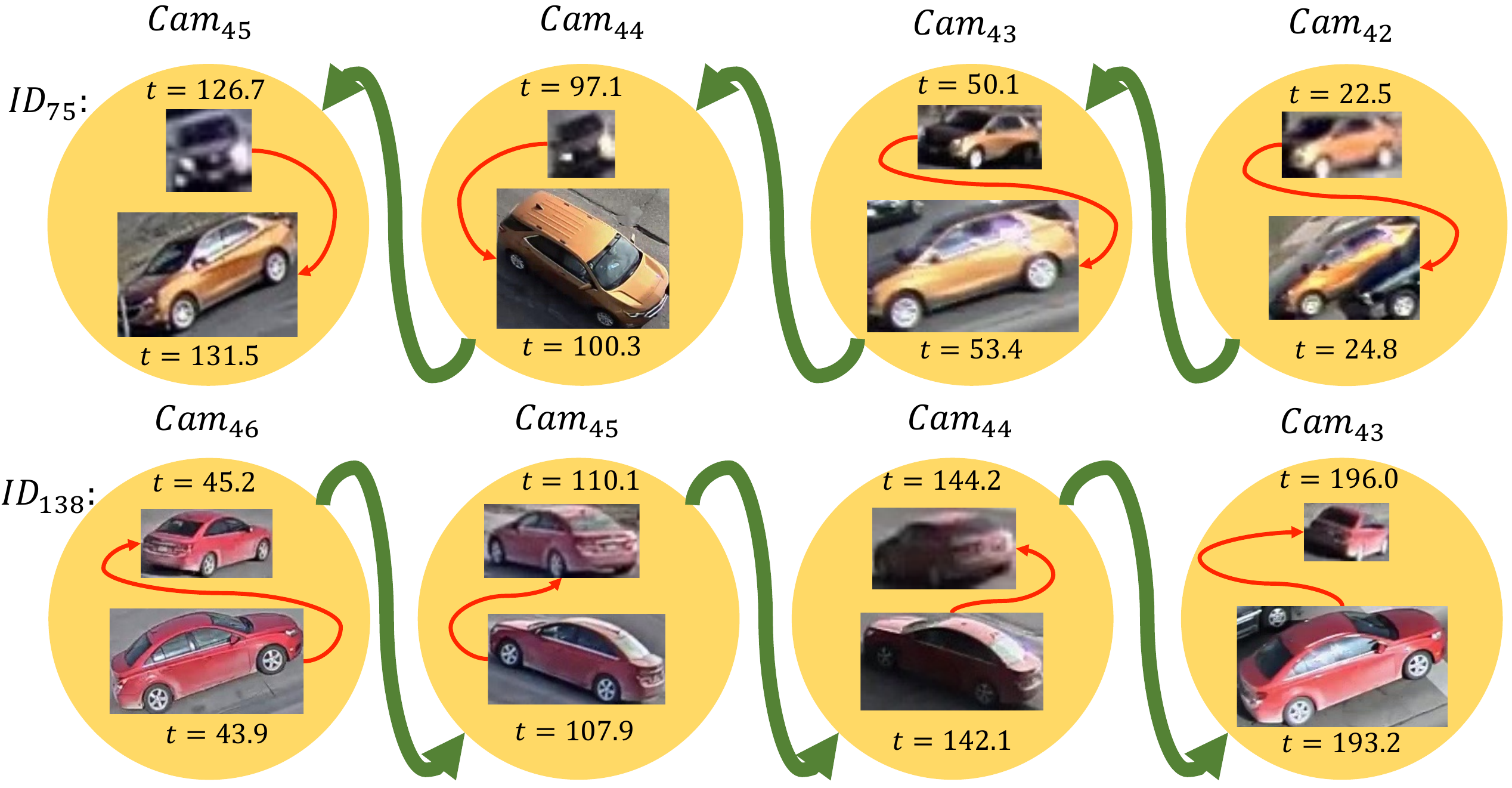}
    \caption{Sample of two Multi Camera tracks generated by our method for the AI City challenge test data.}
    \label{fig:multi_cam_tracking_res_AIC}
\end{figure}
\section{Conclusion}\label{sec:conclusion}
In this work, we highlighted the importance of designing a scalable and real-time multi-camera vehicle tracking system that can provide precise and timely information for transportation applications. Moreover, we shed light on the practical issues involved in achieving this goal and subsequently propose our multi-camera tracking system to address these issues and can process camera feeds in parallel and identify vehicle identities over a network of traffic cameras in real-time. Thanks to its effectiveness, it has been adopted in the Regional Integrated Transportation Information platform to provide real-time understanding of the status of traffic and identify sources of traffic congestion. In addition, we have participated in the 2021 NVIDIA AI City city-scale multi-camera tracking challenge and our model is ranked among the top five contestants without any consideration of processing time and computational complexity of the submissions.

% or

%\input{appendices}


%\input{acknowledgement}

\ifCLASSOPTIONcaptionsoff
  \newpage
\fi

\begin{IEEEbiography}[{\includegraphics[width=1in]{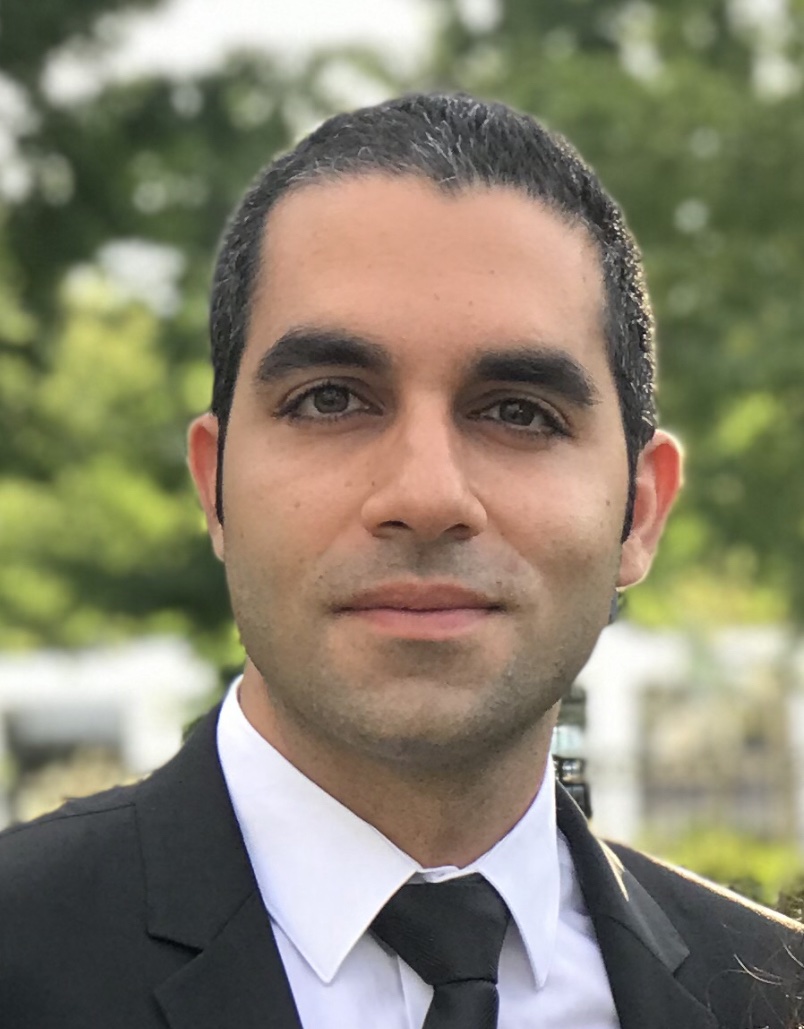}}]{Pirazh Khorramshahi} is a member of Artificial Intelligence for Engineering and Medicine Lab and a Ph.D. candidate in the Electrical and Computer Engineering department at Johns Hopkins University (JHU). Prior to joining JHU in 2020, he was a Ph.D. student in the Electrical and Computer Engineering department at University of Maryland. He also received the B.Sc. and M.Sc. degrees in electrical engineering from the department of Electrical Engineering at Sharif University of Technology in 2013, and 2015 respectively. His general research interests are in Computer Vision and Machine Learning. His Ph.D. research is focused on learning supervised and self-supervised attention models for the task of Vehicle Re-Identification, and he has published papers in International Conference on Computer Vision (ICCV), European Conference on Computer Vision (ECCV), and Computer Vision and Pattern Recognition Workshops (CVPRW). His 2019 ICCV paper on Vehicle Re-Identification was among 4.3\% of 4303 total submissions which was accepted for spotlight oral talk.
\end{IEEEbiography}
\vskip -2\baselineskip plus -1fil
\begin{IEEEbiography}[{\includegraphics[width=1in]{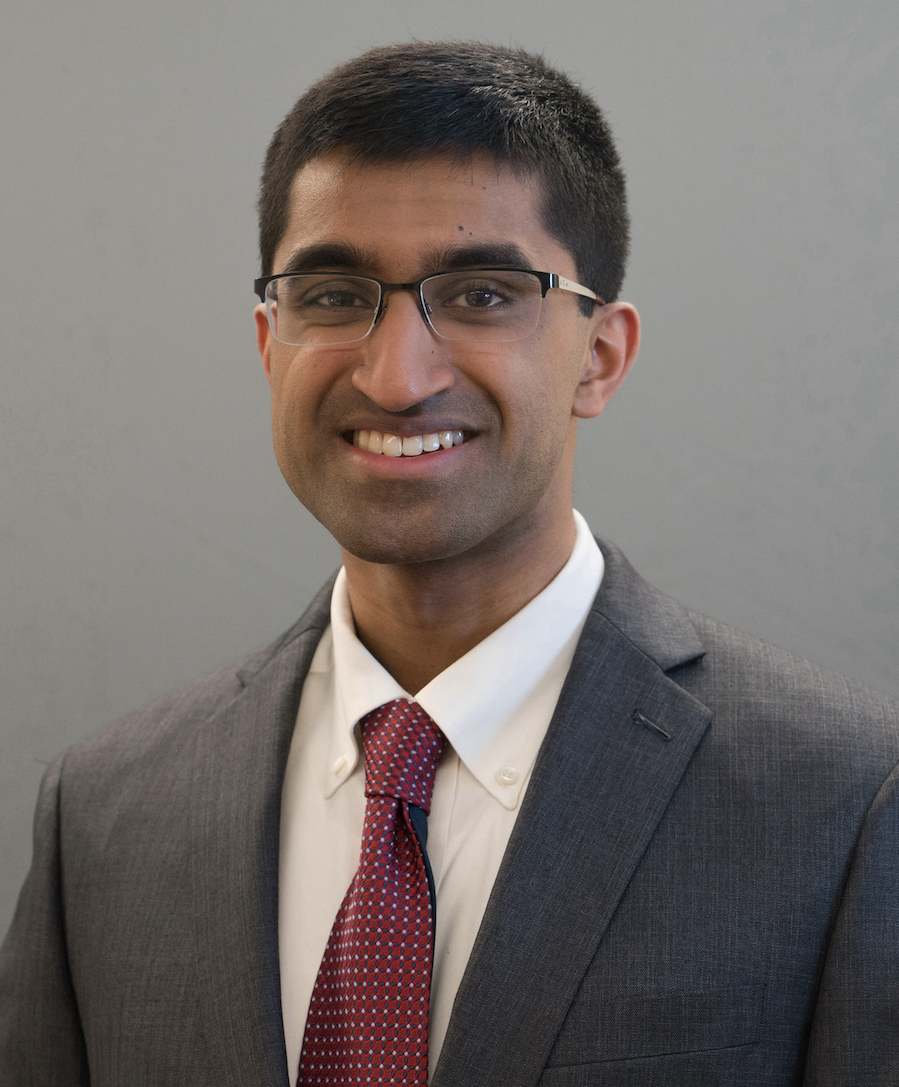}}]
{Vineet Shenoy}is a PhD student in the Department of Electrical and Computer Engineering at Johns Hopkins University. He is broadly interested in Computer Vision and Machine Learning, with a current focus on object detection using domain adaptation and few-shot learning techniques. He has performed research in Spectrum Sharing at the National Institute of Standards and Technology (NIST), and Biometrics at the MITRE Corporation. Previously, he earned his B.S. in Electrical and Computer Engineering and Computer Science at Rutgers University New-Brunswick, where he received the James Leroy Potter Award for Original Investigation as well as the Chancellor's Leadership Award. He is a member of the Phi Beta Kappa and Tau Beta Pi honor societies.
\end{IEEEbiography}
\vskip -2\baselineskip plus -1fil
\begin{IEEEbiography}[{\includegraphics[width=1in]{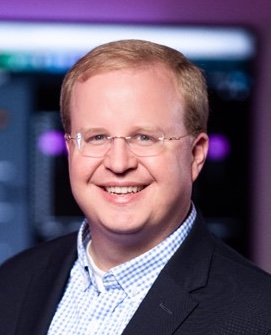}}]{Michael Pack} is the founder and director of the CATT Laboratory where he sets the strategic vision for his team of 100+ data scientists, software developers, systems integrators, and transportation professionals to make federal, state, and local transportation data easily accessible and usable by diverse user communities.  Through the development of innovative applications and data visualization tools, Michael’s team enables data sharing, informed decision-making, better response to emergencies, insight discovery, and increased productivity.  He has been honored at the White House as a Champion of Change for his leadership in integrating vast amounts of data from around the country into the Regional Integrated Transportation Information System, for his performance management applications, and for his efforts in breaking down the barriers within agencies that prevent data from being leveraged to its fullest potential. He has previously worked at the Oak Ridge National Laboratory’s Center for Transportation Analysis and the University of Virginia's Smart Travel Laboratory.
\end{IEEEbiography}
\vskip -2\baselineskip plus -1fil
\begin{IEEEbiography}[{\includegraphics[width=1in]{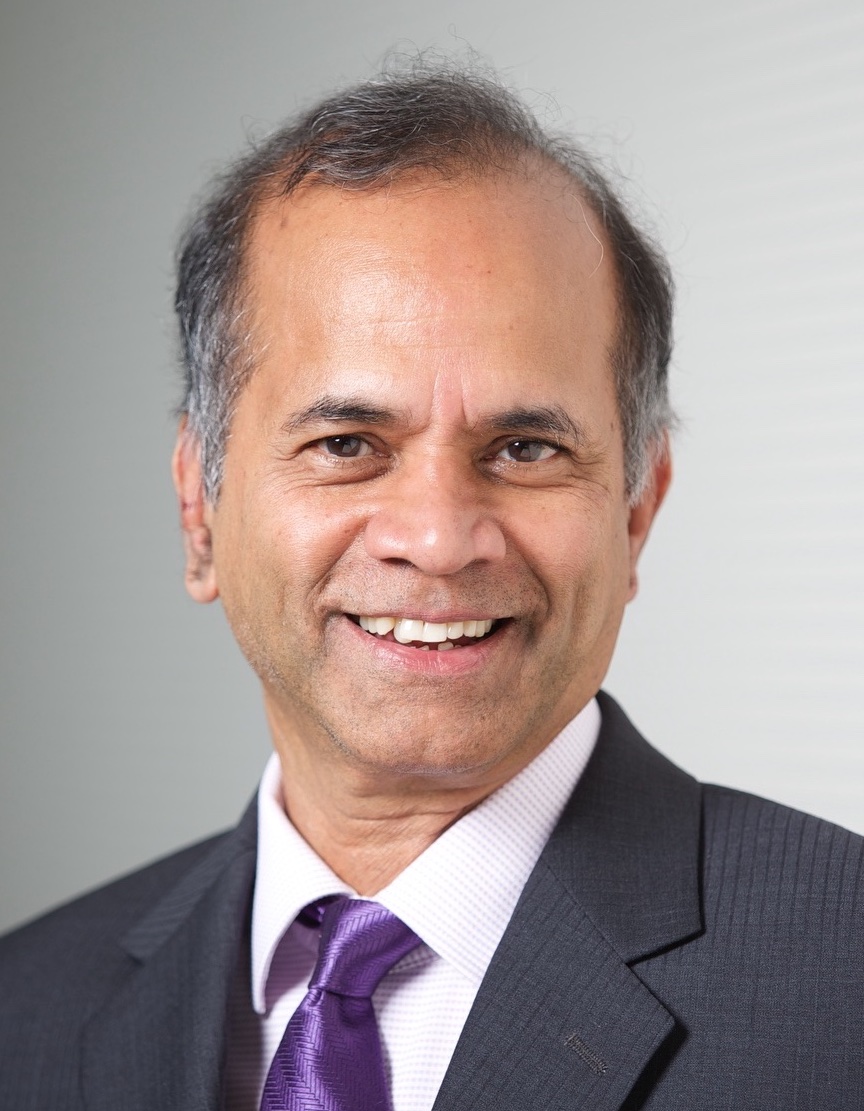}}]{Rama Chellappa} is a Bloomberg Distinguished Professor in the Departments of Electrical and Computer Engineering (Whiting School of Engineering) and Biomedical Engineering (School of Medicine) with a secondary appointment in the Department of Computer Science at Johns
Hopkins University (JHU). At JHU, he is also affiliated with CIS, CLSP, IAA, Malone Center, and MINDS. Before coming to JHU in August 2020, he was a Distinguished University Professor, a Minta Martin Professor of Engineering, and a Professor in the ECE department and a Permanent Member at the University of Maryland Institute Advanced Computer Studies at the University of Maryland (UMD). He holds a non-tenure position as a College Park Professor in the ECE department at UMD. His current researcher interests are computer vision, pattern recognition, machine intelligence and artificial intelligence. He received the K. S. Fu Prize from the
International Association of Pattern Recognition (IAPR). He is a recipient of the Society, Technical Achievement, and Meritorious Service Awards from the IEEE Signal Processing Society and four IBM Faculty Development Awards. He also received the Technical Achievement and Meritorious Service Awards from the IEEE Computer Society. He received the Inaugural Leadership Award from the IEEE Biometrics Council and the 2020 IEEE Jack S. Kilby Medal for Signal Processing. At UMD, he received college and university level recognitions for research, teaching, innovation, and mentoring of undergraduate students. He has been recognized as an
Outstanding ECE by Purdue University and as a Distinguished Alumni by the Indian Institute of Science, India. He served as the Editor-in-Chief of PAMI. He is a Golden Core Member of the IEEE Computer Society, served as a Distinguished Lecturer of the IEEE Signal Processing Society and as the President of the IEEE Biometrics Council. He is a Fellow of AAAI, AAAS, ACM, IAPR, IEEE, NAI, and OSA and holds eight patents.
\end{IEEEbiography}

%\newpage


%\vfill


\end{document}